\documentclass{article}
\usepackage{amsmath}
\usepackage[final, eandd]{neurips_2026}

\newif\ifarxiv
\arxivtrue

\usepackage[utf8]{inputenc}
\usepackage[T1]{fontenc}
\usepackage{hyperref}
\usepackage{url}
\usepackage{booktabs}
\usepackage{amsfonts}

\usepackage{nicefrac}
\usepackage{microtype}
\usepackage{xcolor}
\usepackage{graphicx}
\usepackage{multirow}
\usepackage{array}
\usepackage{tabularx}
\usepackage{tikz}
\usetikzlibrary{positioning, calc, arrows.meta, shapes.geometric}

\newcommand{\rev}[1]{#1}
\newcommand{\revcolor}{}

\title{CogArena: Benchmarking Multimodal Agents on Interactive Behavioral Experiments}

\author{
  Kiant\'{e} Fernandez \\
  Department of Psychology \\
  University of California, Los Angeles \\
  \texttt{kiante@ucla.edu}
  \And
  Caitlin Chen \\
  Department of Psychology \\
  University of California, Los Angeles
  \And
  Jialu Zhou \\
  Department of Psychology \\
  University of California, Los Angeles
\And
  Bartek Sadowski \\
  Department of Psychology \\
  University of California, Los Angeles
  \And
  Anthony C.\ Miceli \\
  Department of Psychology \\
  University of California, Los Angeles
  \And
  Ian Krajbich \\
  Department of Psychology \\
  University of California, Los Angeles \\
  \texttt{krajbich@ucla.edu}
}

\begin{document}

\maketitle

\begin{abstract}
Claims about the cognitive capabilities of large language models are typically tested by translating behavioral paradigms into text prompts, bypassing the visual interfaces and time pressure through which human participants actually engage with experiments. At the same time, AI agents can now complete online behavioral tasks and produce human-like data, which raises concerns that online datasets in the social and behavioral sciences are increasingly open to contamination. We introduce CogArena, a benchmark of ten experimental paradigms from the social and behavioral sciences. Agents complete the experiments through a standard web browser, working with the same visual stimuli and response interfaces that online participants use. A three-level scoring procedure evaluates whether the agent completed the task, how accurately it performed, and whether its data show the behavioral signatures each paradigm reliably produces in people. Across four multimodal models, each run ten times per task, we find that finishing an experiment and reproducing its signatures are different things. Many sessions did not finish, and those that did reproduced some signatures reliably and others not at all. CogArena provides the tasks, scoring code, and a public leaderboard for new submissions.
\end{abstract}

\section{Introduction}
\label{sec:intro}
Large language models are increasingly evaluated on tasks drawn from the social and behavioral sciences, with results used to characterize the models' decision-making, reasoning, and learning capacities~\citep{binz2023using, codaforno2024cogbench, binz2024turning}. These evaluations typically translate behavioral paradigms into natural language prompts. The stimuli, instructions, and response options are rendered as text, and the model's completions are scored against human behavioral data. While experimentally clean, this approach bypasses the interfaces human participants actually use. When people complete experiments online, they navigate web-based platforms, encode visual stimuli, and respond under time pressure with keypresses or mouse movements. Consider a standard interference task such as the Stroop task. Presented as text alone (``the word RED printed in blue ink, press the key corresponding to the ink color''), the task tests language comprehension and instruction following. Presented instead as a colored word on screen, with a two-second response deadline and a fixation cross between trials, the task additionally recruits visual perception, selective attention, and response inhibition. The two versions may be testing different things, and neither necessarily includes the other.

At the same time, AI agents that can complete tasks on the web without supervision have improved quickly. Recent benchmarks show that they can navigate websites, fill out forms, and carry out multi-step tasks through a browser~\citep{zhou2024webarena, yao2022webshop, deng2024mind2web, xu2025turkingbench}. For researchers who collect behavioral data online, this is a practical problem. An agent can sign up for a study, pass the attention checks, and submit answers that look like a person's, and current methods for detecting such respondents do not catch it~\citep{westwood2025threat,fernandez2026threat}. There is also early evidence that agents can reproduce finer-grained signatures of human behavior, such as the shape of a response-time distribution~\citep{vanderstigchel2026online}. What has not been tested is how current models fare with a behavioral experiment as a participant meets it, with a page to operate, stimuli to encode, a deadline on each response, and trials whose meaning depends on the trials before them.

Here we introduce CogArena, a benchmark designed to evaluate how AI agents perform behavioral experiments as humans commonly encounter them in online research settings. Rather than scoring answers to text prompts or annotation tasks, CogArena evaluates whether an agent's behavioral data exhibit the same signatures that humans reliably produce. The benchmark serves two purposes. For researchers studying machine cognition, it tests whether an agent's behavior on well-studied paradigms resembles human behavior beyond getting the answers right~\citep{shiffrin2023probing, hagendorff2023machine, dillion2023can}. For researchers who rely on online data collection, it characterizes which response modalities and signatures agents can and cannot reproduce, and so provides a continuously updated picture of the threats autonomous agents pose to the validity of online behavioral experiments~\citep{westwood2025threat, vanderstigchel2026online,zhang2026estimating}.

This paper makes three contributions. First, we provide an initial release of a curated ten-task benchmark of behavioral experiments spanning learning, risk, social and economic decision-making, memory recall, memory recognition, perception, judgment, foraging, compositionality, and moral decision-making. Each task is adapted from prior work, with parameters and stimuli drawn from published studies and paired with human baselines computed from the source studies' released participant data wherever those data exist. Second, we introduce a three-level scoring procedure that tests whether the agent completed the task, how accurately it performed relative to those baselines, and whether its data show each paradigm's behavioral signatures, with every score computed by a fixed statistical test rather than by a language model. We additionally report a chance floor, the score each level assigns to an agent that responds at random, as a reference against which reported scores can be read. To our knowledge, no prior benchmark scores cognitive signatures in interactive, browser-based agent evaluation. Third, as a preliminary evaluation, we report results from four multimodal models, each driven through the browser by the same software (the Browser-Use agent, which converts a model's output into clicks and keypresses), with ten sessions per model and task. \rev{The central finding is that completing an experiment does not guarantee reproducing the signatures that decades of human research have established as defining features of these tasks. Sessions that reach the end of an experiment reproduce some signatures and miss others, the pattern varies sharply across tasks and models, and many sessions never reach the end at all. Section~\ref{sec:study_gap} gives the evidence for this claim and Section~\ref{sec:limitations} its limits.}\footnote{Code, metadata, and a reproduction recipe are at \url{https://github.com/kiante-fernandez/CogArena} (tagged release \rev{\texttt{v1.3.0}}). The deployed platform and live leaderboard are at \url{https://cog-arena.vercel.app}.}

\section{Related Work}
\label{sec:related}
CogArena draws on two streams of prior work, benchmarks that evaluate AI agents on interactive web-based tasks and assessments of machine intelligence that use cognitive-science paradigms as evaluation instruments.

\paragraph{Interactive agent benchmarks.} Browser-based agent evaluation has progressed from simulated environments~\citep{yao2022webshop} to self-hosted web applications~\citep{zhou2024webarena}, live websites~\citep{deng2024mind2web}, and tasks that take many steps in a graphical interface~\citep{liu2025verigui, li2026agencybench, wei2025evomemory}. Alongside this expansion of task domains, \rev{researchers have argued for} grounding evaluation in verifiable task state rather than in approximate scoring that depends on auxiliary judgments or surface-level cues. Two recent benchmarks illustrate this shift along different axes. GameWorld~\citep{ouyang2026gameworld} grounds scoring in serialized in-game state from browser-based games. HAL~\citep{kapoor2025hal} separates the benchmark (what to measure) from the harness (how an agent connects), a separation CogArena adopts. Most relevant to CogArena's interface is TurkingBench~\citep{xu2025turkingbench}, which shows that the web pages on which crowdworkers complete tasks provide a natural browser-based testbed for agent evaluation. CogArena combines these two design principles, instrumented state scoring and benchmark--harness separation, and extends them to a new kind of content, canonical experimental paradigms with decades of published human data. Performance is evaluated against literature-derived behavioral signatures rather than annotation accuracy or game progress (Table~\ref{tab:comparison}).

\paragraph{The application of cognitive science to artificial systems.} A complementary line of research evaluates whether AI systems exhibit human-like cognitive patterns~\citep{shiffrin2023probing, hagendorff2023machine, aher2023using, dillion2023can}. \citet{binz2023using} applied canonical cognitive paradigms to GPT-3 through text-based prompting, identifying signatures consistent with model-based reinforcement learning alongside failures in causal reasoning. Follow-up work, CogBench~\citep{codaforno2024cogbench}, formalized this approach into a systematic framework, testing 35 models across seven experiments through a text API. Subsequent work, including the Psych-101~\citep{binz2025centaur} and Psych-201~\citep{binz2026posttraining} datasets, has aggregated trial-by-trial human behavioral responses across more than 160 experiments, providing large-scale training data for behavioral foundation models. \rev{All} of these approaches present stimuli in text-based form. In contrast, CogArena evaluates agents in interactive online behavioral experiments, under the stimulus formats, timing constraints, and response requirements under which human behavioral data are typically collected. \begin{table}[t]
\caption{Comparison of CogArena with related benchmarks. CogArena combines a browser-based interface with behavioral experiments scored against literature-derived signatures.}
\label{tab:comparison}
\centering
\small
\begin{tabular}{@{}lllll@{}}
\toprule
\textbf{Benchmark} & \textbf{Content} & \textbf{Interface} & \textbf{Grading} & \textbf{Human reference} \\
\midrule
WebShop & E-commerce & Browser & End-state success & -- \\
WebArena & Web productivity & Browser & End-state (binary) & -- \\
Mind2Web & 137 websites & Browser & Step + end-state & -- \\
TurkingBench & NLP annotation & Browser & Per-trial accuracy & Crowdworkers \\
CogBench & 7 cog.\ experiments & Text API & Cognitive metrics & Qualitative \\
Psych-101/201 & 160+ experiments & Text API & Choice prediction & Trial-by-trial \\
GameWorld & 34 video games & Browser & Success + progress & Novice-player baseline \\
\midrule
\textbf{CogArena} & \textbf{10 cog.\ experiments} & \textbf{Browser} & \textbf{3-level + signatures} & \textbf{Source-study data} \\
\bottomrule
\end{tabular}
\end{table}
\section{The CogArena Benchmark}
\label{sec:benchmark}

CogArena version~1 (v1) consists of ten experiments spanning learning, risk, social and economic decision-making, memory recall, memory recognition, perception, judgment, foraging, compositionality, and moral decision-making (Table~\ref{tab:tasks}). Trial counts range from 16 to 200 per task, and response modalities include keypresses, button clicks, and typed responses. All tasks are implemented as standalone web applications using jsPsych~v8~\citep{deleeuw2015jspsych, deleeuw2023jspsych}, a JavaScript framework for browser-based behavioral experiments, and include instruction screens, practice trials, and main trial blocks. Each task is adapted from a vetted research codebase, with original stimuli and trial-structure conventions retained wherever the source license permits. Per-task configurations and provenance are recorded in Appendices~\ref{app:tasks} and~\ref{app:provenance}\rev{, including, where relevant, a task's correspondence to a Psych-101/201 entry}.

\begin{table}[t]
\caption{The CogArena v1 task set. The last column lists examples of the scored signatures; Appendix~\ref{app:rep_signatures} gives the complete list. Tasks 1--4, 7, and 8 each also carry one response-time relational signature (Appendix~\ref{app:rep_signatures}). See Appendix~\ref{app:tasks} for full per-task scoring specifications and source-version provenance.}
\label{tab:tasks}
\centering
\footnotesize
\resizebox{\textwidth}{!}{%
\begin{tabular}{@{}lllll@{}}
\toprule
\textbf{\#} & \textbf{Category} & \textbf{Task} & \textbf{Source} & \textbf{Scored Signatures} \\
\midrule
1 & Learning & Grid bandit (safe vs.\ risky blocks) & \citep{wu2018generalization, witte2024exploring} & Kraken avoidance, value-directed exploitation, within-block learning \\
2 & Risk & Marbles (developing risk) & \citep{ciranka2022uncertainty} & EV sensitivity, EV-optimal choice, magnitude effect \\
3 & Social / Economic & Repeated PD + Battle of Sexes & \citep{akata2025playing, cooper1989communication} & Tit-for-tat reciprocity, conditional cooperation, coordination \\
4 & Memory recall & Serial recall & \citep{haridi2026context} & Similarity advantage, primacy \\
5 & Memory recognition & Old/new visual recognition & \citep{brady2008visual} & Hit rate above false-alarm rate, low false-alarm rate \\
6 & Perception & Random dot motion & \citep{roitman2002response} & Coherence-dependent accuracy and RT \\
7 & Judgment & Phishing detection & \citep{singh2019training} & Hit rate above false-alarm rate, learning effect \\
8 & Foraging & Effort foraging & \citep{constantino2015learning, bustamante2023effort} & Travel-cost residence effect (MVT), MVT-optimal leaving \\
9 & Compositionality & Tiny Alchemy & \citep{brandle2023empowerment} & Above-chance discovery, non-redundant exploration \\
10 & Moral decision-making & Moral Machine vignettes & \citep{awad2018moral} & Save more, young, human, and lawful \\
\bottomrule
\end{tabular}%
}
\end{table}

\subsection{Task Design Principles}
\label{sec:design}

Unlike prior AI agent benchmarks, which evaluate agents on more general web tasks, CogArena tasks are single-page, multi-trial experiments requiring sustained interaction with 16 to 200 sequential stimuli and responses under configurable per-trial deadlines. This setting reproduces the demands each paradigm places on online participants.

For tasks with randomized stimuli, a per-session seed generates a reproducible stimulus sequence that differs across sessions. The task code is public, so an agent could in principle have seen it, but there is no fixed item set to memorize. Future releases can also rotate the scored set through the forty-task catalog already hosted, each under its own version tag. Even if an agent has encountered text-based descriptions of these paradigms, completing a visual experiment through a browser requires \rev{operating a rendered interface} and executing responses under whatever deadlines the task imposes, capabilities that text-based training may not fully exercise.

\rev{To check that agents actually use what they see, we also re-ran four tasks with the screenshots removed, so that the agent received only the page's underlying text and list of elements (the page's DOM-derived text and list of interactive elements) rather than the rendered page. Composite scores fell by $31$--$53\%$ across the four tasks, significantly on three (Appendix~\ref{app:harness}).}

\subsection{Three-Level Scoring Pipeline}
\label{sec:scoring}

A human participant completing the Stroop task does not merely press the correct keys. They also respond more slowly on incongruent than on congruent trials, known as Stroop interference~\citep{stroop1935studies}, slow down after an error, known as post-error slowing~\citep{dutilh2012testing}, and show less interference after incongruent trials, the congruency sequence or Gratton effect~\citep{gratton1992optimizing}. Accuracy alone misses these signatures. CogArena therefore scores agents at three levels, each more specific than the last. Level~1 asks whether the agent completes the experiment, Level~2 whether it performs accurately, and Level~3 whether its trial data show the signatures that the human literature has established for the paradigm.

Each level produces a score in $[0, 1]$, and the composite CogArena score is a weighted sum scaled to a 0--100 range,
\begin{equation}
\text{CogArena Score} = \bigl(w_1 \, S_{\text{L1}} + w_2 \, S_{\text{L2}} + w_3 \, S_{\text{L3}}\bigr) \times 100,
\qquad w_i \in [0,1],\ \sum_i w_i = 1
\label{eq:composite}
\end{equation}
where $S_{\text{L1}}$, $S_{\text{L2}}$, and $S_{\text{L3}}$ denote the Level~1, Level~2, and Level~3 scores for a given task. \rev{Two tasks have no Level~2 score, for reasons given below. For those, $w_2$ is redistributed to $w_1$ and $w_3$ in proportion, so every task's composite lies on the same 0--100 scale.} A model's overall score at each level is the mean of its per-task scores\rev{, taken over the tasks on which that level is defined}. The weights $(w_1, w_2, w_3)$ are a design choice. We adopt one scheme for the headline results and report sensitivity to alternatives.

\rev{If a task returns no scores at all, for example because the agent crashed before submitting any trials, the leaderboard counts it as $0$ in the model's mean. Averaging only over the tasks that returned scores would reward a model for abandoning the tasks it does poorly on. Every session in this study returned scores, so this rule did not apply here. Missing data within a session are handled at Level~3, as described below. The composite is a convenient summary for ranking, but the per-level breakdown is the primary reporting unit.}

\paragraph{Level~1: Task Completion.}
The completion level checks that the agent engaged with the experiment. Five binary checks assess whether trial data were received, whether enough trials were completed (at least 80\% of those expected), whether the timeout rate was below 20\%, whether responses used valid inputs, and whether the experiment reached its end. The Level~1 score is the proportion of these checks that passed.

\paragraph{Level~2: Performance Accuracy.}
The accuracy level compares the agent with the participants of each task's source study on a set of task-specific metrics (Appendix~\ref{app:provenance}). Metrics include proportion correct, mean reward, and hit and false-alarm rates~\citep{green1966sdt}. \rev{For each metric $m$, the scorer computes the agent's value $x_m$ from its trial data and scores it as $\Phi\bigl((x_m - \mu_m)/\sigma_m\bigr)$, where $\mu_m$ and $\sigma_m$ are the human mean and standard deviation and $\Phi$ is the standard normal CDF. A score of $0.5$ therefore means the agent matched the average participant, and the score approaches $1$ as the agent exceeds the human mean by more and more. For metrics on which lower is better, such as false-alarm rates, the sign of the difference is reversed. Because the score keeps rising as the agent outperforms people, Level~2 holds only metrics on which exceeding the human value is better performance. A preference has no better direction, so scoring one here would reward an agent for being more extreme than people. Preferences are tested at Level~3 instead, which asks only whether the agent shares their direction. The \texttt{moral\_machine} task records only preferences and \texttt{marbles\_risk} only the rate of higher-expected-value choices, which treats risk neutrality as correct, so neither task has a Level~2 score. For the same reason, \texttt{effort\_foraging} does not score its stay rate or residence times, since staying longer is better only up to the optimum the marginal value theorem sets (Appendix~\ref{app:metrics}).} The Level~2 score is the mean of the metric scores for the task. Across the v1 set the benchmark computes \rev{31} accuracy metrics over \rev{eight} tasks, \rev{3.9} per task on average (per-task counts in Appendix~\ref{app:metrics}). \rev{Absolute response time is not among them. It is scored instead through within-session relational signatures at Level~3, for reasons given in Section~\ref{sec:discussion}.}

\paragraph{Level~3: Behavioral Alignment.}
The behavioral alignment level tests whether agent trial data show the cognitive signatures that characterize human performance on each paradigm (e.g., coherence-dependent accuracy in random dot motion or tit-for-tat reciprocity in repeated games). Each task specifies a set of signatures, and each signature defines a statistical test, an expected direction, a significance threshold ($p < 0.05$), and an importance weight $w_j$. Appendix~\ref{app:tests} describes the four test types, which are exact one-sample proportion tests, paired proportion contrasts, conditional proportion contrasts, and Pearson or Spearman correlations. A signature scores 1.0 if the effect is significant in the expected direction, 0.5 if it is in the expected direction but not significant, and 0.0 otherwise. The Level~3 score is the weighted mean
\begin{equation}
S_{\text{L3}} = \frac{\sum_{j \in \mathcal{T}} w_j \, s_j}{\sum_{j \in \mathcal{T}} w_j}, \qquad s_j \in \{0, 0.5, 1\},
\label{eq:l3}
\end{equation}
where $\mathcal{T}$ is the set of signatures that could be tested in that session (defined below), out of the $n_k$ signatures specified for task $k$, and $w_j$ reflects how central the signature is to the task's primary phenomenon. Across the v1 set the benchmark evaluates \rev{39} signatures, \rev{3.9} per task on average (full specifications in Appendix~\ref{app:signatures}). Because signatures are tested independently, some may reach significance by chance. We do not adjust for multiple comparisons, because each pre-specified signature is reported on its own rather than as part of a single confirmatory family, and the chance floor described below gives an empirical reference for the score such testing produces under random behavior. Trials within an agent's session are also serially dependent. In the reported scores, proportion tests run at an effective sample size and paired contrasts take block-bootstrap $p$-values, whereas correlation tests and the conditional contrasts use their ordinary $p$-values; Appendix~\ref{app:power} re-runs every test with a block bootstrap as a sensitivity check (Appendix~\ref{app:tests}).

\rev{A low Level~3 score can arise in three ways. The agent may genuinely not show the signature; the session may have ended too early, or with too little variation in responses, for the signature to be tested; or the model may never have seen the information the signature depends on (Section~\ref{sec:design}). Only the first says anything about the agent, so the scoring must keep the other two from counting as failures. A session that ends early, or an agent that gives the same response throughout, can leave too few trials in the relevant condition to test. Scoring such a signature $0$ would treat a missing measurement as an absent effect, so it is left out of Equation~\ref{eq:l3}, and Level~3 is the weighted mean over the signatures that could be tested. A session in which no signature could be tested ($\mathcal{T}$ empty) receives a Level~3 of $0$. Appendix~\ref{app:power} reports how many signature observations were untestable, and Table~\ref{tab:sig_pass} gives the testable count for every signature.}

\rev{A Level~3 computed from one testable signature out of five says little about the session. We therefore call a session measurable when at least half of its task's signature weight is testable and no test errored, and we report averages in two ways. Averages over all sessions count every session as scored, and are used for ranking and on the leaderboard. Averages over measurable sessions describe behavior where there was enough data to measure it. Section~\ref{sec:study} reports model rankings under both.}

For all results in this paper we used $(w_1, w_2, w_3) = (0.15, 0.35, 0.50)$. \rev{Completion is a prerequisite and takes the lowest weight, accuracy captures conventional performance, and behavioral alignment is the distinctive contribution and takes the highest weight.} Appendix~\ref{app:weights} reports sensitivity analyses under three alternative weightings and three alternative Level~3 grading rules.

Lastly, to anchor the interpretation of all three levels under random responding, we computed a chance floor. We ran a Random Agent, which presses a random valid key on each trial, ten times per task and scored its \rev{one hundred} sessions with the same code (Appendix~\ref{app:chance}). We report this chance floor as a reference column in results tables and as a reference line in figures.

\section{Methods}
\label{sec:methods}

\subsection{Platform}
CogArena provides the experiments and the scoring, but not the agent (i.e., the harness). The software that lets a model operate a browser, which we call the scaffold, is supplied by whoever submits to the benchmark. A scaffold takes the model's text output and turns it into clicks, keypresses, and page navigation, and passes screenshots or page text back to the model. Examples include Anthropic's and OpenAI's Computer Use tools, Browser-Use~\citep{browseruse2024}, OpenHands~\citep{wang2024openhands}, and custom code written with a browser-automation library such as Playwright. The platform itself provides the tasks, an interface for starting sessions and submitting trial data, the scoring pipeline, and a public leaderboard.

\begin{figure}[t]
\centering
\includegraphics[width=\textwidth]{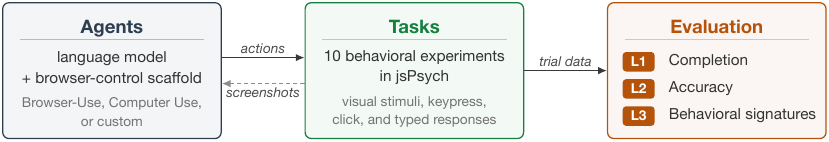}
\caption{CogArena conceptual architecture. An agent (using any scaffold) interacts with CogArena tasks (behavioral experiments served by the platform), and the resulting trial data are scored at three levels, L1 task completion, L2 performance accuracy, and L3 alignment with task-specific signatures. Trial data are sent to the server during the session, so partial sessions still produce usable data.}
\label{fig:architecture}
\end{figure}

The platform is a web server that hosts the ten jsPsych experiments and accepts requests from agents to start a session, submit trial data, and score the session (Figure~\ref{fig:architecture}). An agent starts a session, opens each task's web page in its browser, and completes the experiment as a participant would. Once every task in a session has been received, the server scores it. All experimental logic (stimulus presentation, response recording, timing, and condition randomization) lives in jsPsych, so the server only stores and scores what the page sends.

The platform does not constrain how an agent controls the browser. The model may see the page as a screenshot, as the underlying page text, or both, and any scaffold may be used. When a session is created, the agent reports which scaffold it uses and what the model sees, and these are stored as metadata that can be used to compare models. \rev{The setup used in the present study is described in Section~\ref{sec:study_design} and in full in Appendix~\ref{app:harness}.} A plain-text instruction file (i.e., a SKILL.md) explains how to start a session and complete each task. Response deadlines can be modified for a session, and by default jsPsych will record the response time whether or not a deadline is in force. The platform is a public website on which agents complete the same experiments under the same conditions, and every scored session appears on a shared leaderboard ranked by composite score.

\subsection{Reference Agents and Public Submissions}
We provide two agents with the platform. The first, the Random Agent, has no language model at all. Instead, it guesses by pressing a randomly selected valid key on each trial. It serves two purposes. It verifies that the experiments and scoring run end to end, and it establishes the chance floor described in Section~\ref{sec:scoring}. The second, the Browser-Use Agent, is the scaffold used in this study, and it can drive any model available through the major model providers (Section~\ref{sec:study_design}). Anyone can also submit sessions from their own agent, and submissions so far include commercial agent products and several custom scaffolds.

\subsection{Study Design}
\label{sec:study_design}

We evaluated four multimodal models on the ten tasks, all through the same scaffold, Browser-Use~\citep{browseruse2024}. The models were \texttt{kimi-k2.5} (Moonshot AI), \texttt{gemini-3-flash-preview} (Google), \texttt{qwen3-vl-235b-instruct} (Alibaba), and \texttt{gpt-5.4-nano} (OpenAI). All four sit in the inexpensive tier of their providers' offerings, which we refer to as economy-tier models. Model calls were routed through OpenRouter, a service that provides one interface to many model providers. \rev{We define a cell as one model--task combination. Each of the $40$ cells was run ten times, each session with a new random seed for the stimulus sequence, giving $400$ sessions in total. Every session returned Level~1 and Level~3 scores, and Level~2 was defined for eight of the ten tasks. Running each cell ten times lets us report how much scores vary from session to session rather than a single value (Section~\ref{sec:study_gap}).}

\rev{Because all four models are in the economy tier, a low score could indicate either that current multimodal models cannot reproduce a signature or that only less expensive models cannot. To separate these, we also evaluated one state-of-the-art model at the time of writing, \texttt{gpt-5.6-sol}, on each of the ten tasks using the same scaffold. We report it separately from the four-model panel, since a single session per task is too little to compare with ten (Appendix~\ref{app:flagship}).}

\rev{The scaffold determines what the model sees and how its responses reach the page, so our results describe each model together with Browser-Use rather than the model alone. At every step (one model call, producing one or more actions), we recorded whether the model was given a screenshot or only the page text. Every step of all $400$ sessions used screenshots, so every model saw the rendered page throughout. The Random Agent, which has no model, reads the page text only. Each session was limited to $1800$\,s and $500$ steps per task, whatever the task's number of trials (Table~\ref{tab:trialcounts}).}

\rev{Agent sessions were run with the per-trial response deadlines switched off, so that a slow model response could not by itself count as a missed trial.} The Random Agent was run ten times per task under the same scoring, giving the chance floor of Section~\ref{sec:scoring}; all one hundred of its sessions were scored. \rev{The full run settings are listed in Appendix~\ref{app:harness}.} All results in the paper come from one released version of the tasks and scoring code (\texttt{v1.3.0}), and every number can be recomputed from the released results files (Appendix~\ref{app:tasks}).

\section{Results}
\label{sec:study}
\begin{figure}[!htbp]
\centering
\includegraphics[width=\textwidth]{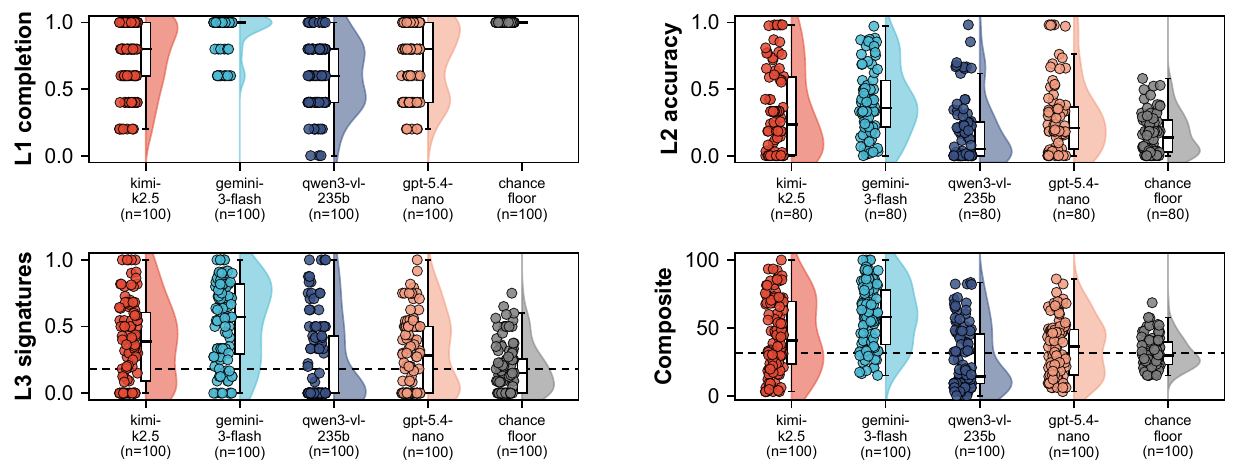}
\caption{Per-model score distributions across the v1 ten-task set, by level and composite. Each panel shows a half-violin, a boxplot, and \rev{one jittered point per scored session. The L3 and Composite panels include every scored session, so a session in which no signature could be tested enters both with a Level~3 of zero.} The dashed horizontal line in those panels marks the random-agent chance-floor mean. The $n$ under each column is the number of \rev{sessions} contributing data, and the right-hand chance-floor column is computed across \rev{all one hundred} random-agent sessions.}
\label{fig:scores}
\end{figure}

\rev{Figure~\ref{fig:scores} shows the distribution of scores for each model at each level. Of the four, \texttt{gemini-3-flash-preview} scored highest on the composite, and this held whichever way sessions were averaged (in $2000$ bootstrap resamples of the ten repeats, it ranked first every time). Its mean composite was $58.6$ over all $100$ sessions and $58.7$ over the $96$ in which Level~3 was measurable. The order of the other three models depended on how sessions were averaged. Over all sessions, \texttt{kimi-k2.5} scored $44.2$, \texttt{gpt-5.4-nano} $36.1$, and \texttt{qwen3-vl-235b-instruct} $26.4$. Over measurable sessions only (those in which at least half of the task's signature weight could be tested), \texttt{kimi-k2.5} scored $50.7$ ($n=76$), \texttt{qwen3-vl-235b-instruct} $48.3$ ($n=41$), and \texttt{gpt-5.4-nano} $45.7$ ($n=63$). The \texttt{qwen3-vl-235b-instruct} model moves up because it finished few of the tasks it started. In the $41$ sessions it did finish well enough to measure, it performed like the rest of the panel. Its failure to finish is recorded separately, at Level~1.}

\subsection{The Completion-vs-Signature Gap}
\label{sec:study_gap}
\rev{We first ask whether the agents finished the experiments. Completion was far from guaranteed. Only $167$ of the $400$ sessions passed all five Level~1 checks, and in $124$ sessions fewer than half the behavioral signatures could be tested, $83$ of them with no testable signature at all. The time and step limits explain little of this. Of the $233$ sessions that failed a Level~1 check, $30$ ran out of time and $1$ ran out of steps, and $11$ were stopped by the scaffold after ten failed actions in a row. The other $191$ ended because the agent declared the task finished when the experiment had not yet ended. The Random Agent, by contrast, completed every session it started (Level~1 $= 1.000$), and no model matched it; the best, \texttt{gemini-3-flash-preview}, reached $0.936$. This is expected. A script that presses a key on every trial should always complete the study. Finishing an experiment is therefore not by itself evidence of ability. What completion does tell us is whether the agent and the platform worked together well enough to produce data.}

\rev{We next ask whether the sessions that produced data showed the signatures. Over the measurable sessions, agents scored well above chance on both remaining levels. Mean Level~2 was $0.345$ against a chance floor of $0.159$, and mean Level~3 was $0.490$ against $0.184$. However, that mean conceals large differences between signatures (Table~\ref{tab:sig_pass}). Some were reproduced in nearly every session that could test them. These were above-chance discrimination of phishing from legitimate email ($90\%$ of testable sessions), the preference for saving more people in the moral dilemmas ($100\%$), and cooperating after the opponent cooperated in the prisoner's dilemma ($100\%$). Others were reproduced in none. These were recognizing previously seen images above chance ($0$ of $25$), staying longer in a patch when travel is costly ($0$ of $32$), and faster responses at higher motion coherence ($0$ of $21$), although for the last of these the scaffold itself is a likely cause, since a single screenshot carries no motion (Appendix~\ref{app:harness}). Completion and signatures came apart in both directions. Agents finished the grid bandit without choosing the tiles their own reward history favored in $44\%$ of sessions, and finished the foraging task without ever showing the travel-cost effect. Even among the sessions that passed every Level~1 check, only $40\%$ of testable signatures were reproduced. Whether an agent completed a task thus says little about whether its data look like a person's.}

\rev{The repeats showed that a single session can be misleading. In $28$ of the $40$ cells, the same model on the same task scored zero on Level~3 in some sessions and above zero in others. Across sessions within a cell, the standard deviation of Level~3 had a median of $0.20$. At the level of signatures, $60$ of the $139$ model--signature combinations passed in some sessions and failed in others. For example, \texttt{kimi-k2.5} on \texttt{marbles\_risk} had a composite of $63.2 \pm 24.8$, so one session could have shown strong sensitivity to expected value or none (Appendix~\ref{app:per_task}). The model ranking was stable under resampling, but the individual cells were not, so our claims are about the panel as a whole.}

\rev{Finally, the state-of-the-art model, \texttt{gpt-5.6-sol}, run once per task, had a higher mean completion score than the panel (Level~1 of $0.960$ against $0.750$, trailing the panel only on serial recall) and scored higher on Level~3 ($0.656$ against $0.490$ over measurable sessions), but beat the best panel model on only four of the ten tasks. With one session per task we cannot say how much of this would survive repetition, so we take it as a sign that a stronger model can do better, not as an estimate of how much better (Appendix~\ref{app:flagship}).}

\subsection{Robustness and Open Leaderboard}
The all-session composite ranking was unchanged under three alternative weightings, $(0.10, 0.30, 0.60)$, $(0.20, 0.40, 0.40)$, and $(0, 0, 1)$. The measurable-session ranking was unchanged under three alternatives to the $\{0, 0.5, 1\}$ Level~3 grade, which shifted the agent--floor gap by at most $0.021$ (Appendix~\ref{app:weights}). New community submissions appear on the shared leaderboard at \url{https://cog-arena.vercel.app/leaderboard}, with per-level columns and the chance-floor reference always shown.

\section{Discussion}
\label{sec:discussion}

We built a benchmark on which AI agents complete ten behavioral experiments through a browser, as online participants do, and scored their data for completion, accuracy, and the behavioral signatures each paradigm reliably produces in people. Four multimodal models, each run ten times per task, often failed to finish, and when they did finish they reproduced some signatures and not others, in a pattern that varied across tasks and across sessions of the same task. An evaluation that scores only whether an agent finished, or only how accurate it was, would not have shown this. The result bears on both audiences for the benchmark. For those asking whether AI systems behave like people~\citep{shiffrin2023probing, hagendorff2023machine, dillion2023can}, it says which human signatures current agents reproduce and which they do not. For those who collect behavioral data online and worry that agents are contaminating it~\citep{westwood2025threat, vanderstigchel2026online,zhang2026estimating}, it suggests which measures may still separate people from current agents.

\rev{A low Level~3 score should not be read as a deficit in ability, because human behavior is not always the best behavior~\citep{ying2025benchmarking}. We therefore labeled every signature as either a capacity or a bias. A capacity is a pattern that a normative model of the task also predicts, so that showing it and performing well are the same thing. For example, an ideal observer's accuracy rises with motion coherence, and a forager that maximizes its rate of reward stays longer in a patch when travel between patches is costly~\citep{charnov1976optimal}. Failing a capacity signature is a performance deficit. A bias, on the other hand, is a pattern that departs from the normative benchmark, or a preference that no benchmark prescribes, so an agent that does not show it may be performing as well as people or better.}

\rev{The repeated prisoner's dilemma illustrates the distinction. The opponent always plays tit-for-tat, so the agent faces a sequential decision problem rather than a game, and over the fifteen rounds shown on screen the policy that earns the most cooperates in every round but the last one. That policy cooperates whenever the opponent has just cooperated, so this signature is a capacity. It also resumes cooperation immediately after a defection rather than retaliating in the next round, so tit-for-tat reciprocity is a bias. In the risky-choice task, expected-utility theory predicts that the risky option will be chosen more often as its probability or payoff rises, so the sensitivity and magnitude signatures are capacities. The theory does not fix how much risk a decision-maker should accept, so choosing the higher-expected-value option, as a risk-neutral decision-maker would, is a preference it permits but does not require. The four moral-dilemma preferences (sparing more people, the young, humans over pets, and the lawful) have no agreed normative standard.}

\rev{The two kinds of signatures gave different results. Over measurable sessions, agents reproduced capacity signatures in $31\%$ of the cases in which they could be tested and bias signatures in $66\%$ (with the response-time signatures discussed below set aside, $36\%$ against $66\%$). The overall Level~3 score is therefore raised by patterns that good performance does not require, and claims about what agents can do should rest on the capacity signatures. For detecting AI-generated data, the same result cuts the other way. The human patterns that current agents reproduce most reliably are preferences, so screening respondents by whether they show the expected preference will not separate agents from people. For example, every agent session that could be tested preferred saving more lives in the moral dilemmas.}

\rev{Response time needs the same care. Agents took up to ten times longer than people to respond (about twelve times, in median response time, on random dot motion), mostly because each response waits for the model to process a screenshot and produce an action. Comparing an agent's response times with human ones therefore measures how fast the model is served rather than how it decides. What can be scored is relative speed within a session, since an agent can be uniformly slow and still respond faster on easy trials than on hard ones, as people do. The seven response-time signatures in the benchmark test relations of this kind, and agents passed them in $30$ of $217$ testable cases. Response times did vary within sessions (median coefficient of variation $0.27$ to $0.58$ across tasks), so the absence of these relations is not because agents responded at a fixed rate. Slow responding may still be a useful cue for flagging AI-generated data in online studies; it is just not a measure of the agent's decision process.}

To conclude, CogArena separately measures whether an agent can complete a behavioral experiment and whether its data reproduce the behavioral signatures seen in people. We hope it will serve as a shared benchmark for two increasingly intertwined communities, AI-agent evaluation and online behavioral research. The tasks, scoring procedures, and leaderboard are open to new submissions, so that its picture of current agents can be updated as models change.
\section{\rev{Limitations}}
\label{sec:limitations}

\rev{First, and most importantly, no human participants have completed our versions of the tasks. Of the $31$ Level~2 baselines, $24$ are computed from participant data released by the original studies (Appendix~\ref{app:provenance}). Because those participants completed the original experiments rather than ours, a difference between agents and people may partly reflect our implementation rather than the agents. This concern is largest for Level~2, which compares magnitudes, and smallest for Level~3, which compares directions. We therefore make no claim about how agents compare with people completing CogArena itself. That comparison requires human data collected on the platform. The platform already serves the same experiments to human visitors, so collecting those data is a matter of recruitment rather than new software, and it is the first priority for the next release.}

\rev{Second, the models we evaluated are a narrow sample. All four are economy-tier, and even with ten sessions per cell their means have wide intervals. The state-of-the-art model was run only once. Every model ran through the same scaffold, Browser-Use, so differences between models are differences between model-and-scaffold combinations, and the screenshot-removal check shows how Browser-Use uses the screenshots rather than how the models themselves could.}

\rev{Third, the current task set does not test every phenomenon these paradigms are known for. The grid bandit, for example, keeps the number of remaining choices fixed, so it cannot test whether an agent explores more when more choices remain. Signatures and manipulations can be added to each task, and the present set should be read as a first release rather than an exhaustive one.}

\begin{ack}
IK thanks the U.S.\ National Science Foundation for support under Award No.~2612963. The authors declare no competing interests.
\end{ack}


\bibliographystyle{plainnat}
\bibliography{references}

\newpage
\appendix
\section{Task Implementation Details}
\label{app:tasks}

\rev{\paragraph{Version.} Every number in this paper comes from one released version of the tasks and scoring code, \texttt{v1.3.0}, applied to the raw trial data of four sets of sessions: the four-model panel at ten sessions per cell, the hundred-session chance floor, the screenshot-removal check, and the single run of the state-of-the-art model. Earlier releases (\texttt{v1.0.0}, \texttt{v1.1.0}, \texttt{v1.1.1}, and \texttt{v1.2.1}) are tagged in the public repository, and no number from them appears in any figure or table here. We state this because a benchmark whose scoring rules change over time can otherwise leave a reader unsure which numbers are comparable.}

\paragraph{Scope.} This paper evaluates the ten curated tasks listed in Table~\ref{tab:tasks} (\texttt{random\_dot\_motion}, \texttt{grid\_bandit}, \texttt{marbles\_risk}, \texttt{repeated\_games}, \texttt{serial\_recall}, \texttt{visual\_recognition}, \texttt{effort\_foraging}, \texttt{tiny\_alchemy}, \texttt{moral\_machine}, \texttt{phishing\_detection}). The deployed CogArena platform additionally hosts a wider community catalog of forty tasks under \texttt{tasks/} at \url{https://github.com/kiante-fernandez/CogArena/tree/v1.3.0/tasks}. Their per-task scoring specifications and trial counts are documented in the public repository but are out of scope for this paper. Throughout the paper, task names drop the version suffix used in the repository: \texttt{random\_dot\_motion}, \texttt{serial\_recall}, and \texttt{phishing\_detection} refer to the released \texttt{random\_dot\_motion\_v2}, \texttt{serial\_recall\_v2}, and \texttt{phishing\_detection\_v2} tasks, which replaced earlier tasks of the same name that remain in the catalog.

\paragraph{File layout.} Each CogArena task is a standalone web application with a consistent file structure. The entry point is an \texttt{index.html} file that loads jsPsych v8 and the experiment script. The experiment logic resides in \texttt{experiment.js}, which defines the full trial sequence including instruction screens, practice blocks with feedback, experimental blocks, and automatic data submission. A \texttt{task\_config.json} file specifies all configurable parameters (number of trials, number of blocks, response keys or slider range, timing parameters, and condition proportions). The scoring directory contains \texttt{level2\_metrics.json} specifying accuracy metrics with human baseline statistics, and \texttt{level3\_signatures.json} specifying behavioral signatures with their statistical tests and expected directions.

\rev{\paragraph{Relation to Psych-101/201.} Those corpora informed which paradigms we selected, but the v1 set is not drawn from them. Three tasks share a source study with a Psych-201 entry. These are \texttt{repeated\_games}~\citep{akata2025playing}, \texttt{moral\_machine}~\citep{awad2018moral}, and \texttt{phishing\_detection}~\citep{singh2019training}. The other seven correspond to no entry in either corpus, which each task records explicitly in its \texttt{task\_config.json}. Even for the three, the correspondence is at the level of the source study rather than the implementation, since our implementations regenerate stimuli from a per-session seed and present them in a browser rather than as transcribed text.}

\paragraph{Stimulus randomization.} Stimulus randomization is seeded from the session identifier, so each evaluation session produces a unique but reproducible stimulus sequence. In \texttt{random\_dot\_motion}, coherence levels are interleaved within each block. In \texttt{grid\_bandit}, the locations of high-value tiles and Kraken tiles are regenerated for each block. In \texttt{marbles\_risk}, gamble values and probabilities are sampled from the task's pre-registered grid. In \texttt{tiny\_alchemy}, the starting set of base elements and its presentation order are fixed. In \texttt{repeated\_games}, the opponent strategy is identical across sessions (tit-for-tat in the prisoner's dilemma (PD) and an alternating opponent in the battle of the sexes (BoS)), but the state on each trial evolves with the agent's actions.

\section{Behavioral Signatures Evaluated in v1}
\label{app:rep_signatures}

Table~\ref{tab:signatures} lists all \rev{39} Level~3 behavioral signatures evaluated across the ten v1 tasks, the statistical test used for each, and the importance weight $w_j$. Per-task literature sources for expected directions are given in Appendix~\ref{app:signatures}. Test types and weights map directly to the Level~3 scoring formula (Equation~\ref{eq:l3}).

\rev{Table~\ref{tab:tasks} lists only scored signatures. Two phenomena these paradigms are known for are deliberately absent. Horizon-dependent exploration in \texttt{grid\_bandit} is not operationalized, because the task has no horizon manipulation. Our version of \texttt{serial\_recall} also does not predict a U-shaped serial-position curve. It inserts a five-second filler interval, the standard manipulation for abolishing recency~\citep{glanzer1966}, so scoring recency would test our implementation rather than the agent. Only primacy is scored.}

\rev{The Type column records each signature's \texttt{construct\_type}, which is stored in the task specification rather than assigned after the fact. C (capacity) marks a signature for which a normative model of the task predicts the same pattern people show, so reproducing it and performing well coincide and failing it is a performance deficit. B (bias) marks a signature that departs from a normative benchmark or expresses a preference that no normative benchmark prescribes, so an agent may reasonably not share it. Thirty-three signatures are capacities and six are biases, namely the four \texttt{moral\_machine} preferences, \texttt{repeated\_games}'s tit-for-tat reciprocity, and \texttt{marbles\_risk}'s above-chance choice of the option with the higher expected value (EV). Each bias signature carries a written rationale in its specification. Over measurable sessions, agents reproduced capacity signatures in $31\%$ of the cases in which they could be tested and bias signatures in $66\%$, so the overall Level~3 mean overstates their abilities, and claims about ability should rest on the capacity signatures (Section~\ref{sec:discussion}). For detecting AI-generated data, both kinds are useful, since a bias that people reliably show separates people from agents whatever its normative status.}

\begin{table}[h]
\caption{Behavioral signatures evaluated in v1, per task. Test types are prop (exact one-sample proportion test), paired (paired proportion contrast), contrast (conditional proportion contrast), and corr (Pearson correlation, or Spearman for the response-time signatures). Appendix~\ref{app:tests} defines each test. \rev{Type is C for capacity (reproducing the human pattern coincides with performing well) or B for bias (reproducing it is not normatively required).}}
\label{tab:signatures}
\centering
\small
\begin{tabular}{@{}llccl@{}}
\toprule
\textbf{Task} & \textbf{Signature} & \textbf{Test} & \textbf{\rev{Type}} & \textbf{$w_j$} \\
\midrule
\multirow{4}{*}{\texttt{random\_dot\_motion}}
  & Coherence-accuracy psychometric & corr   & \rev{C} & 2.0 \\
  & Coherence-RT chronometric        & corr   & \rev{C} & 1.5 \\
  & Above-chance accuracy at low coherence & prop & \rev{C} & 0.5 \\
  & Above-chance overall accuracy   & prop   & \rev{C} & 1.0 \\
\midrule
\multirow{5}{*}{\texttt{grid\_bandit}}
  & Kraken avoidance in risky blocks & prop   & \rev{C} & 1.5 \\
  & Value-directed exploitation      & corr   & \rev{C} & 1.5 \\
  & Above-chance reward              & prop   & \rev{C} & 1.0 \\
  & Within-block learning            & corr   & \rev{C} & 1.0 \\
  & \rev{Within-block RT speed-up}   & \rev{corr} & \rev{C} & \rev{1.0} \\
\midrule
\multirow{4}{*}{\texttt{marbles\_risk}}
  & EV sensitivity                   & corr   & \rev{C} & 2.0 \\
  & Above-chance EV-optimal choice   & prop   & \rev{B} & 1.5 \\
  & Magnitude effect for high payoff & corr   & \rev{C} & 1.0 \\
  & \rev{EV-difference RT effect}    & \rev{corr} & \rev{C} & \rev{1.0} \\
\midrule
\multirow{4}{*}{\texttt{repeated\_games}}
  & Tit-for-tat reciprocity (PD)     & \rev{contrast} & \rev{B} & 1.5 \\
  & Above-chance coordination (BoS)  & prop   & \rev{C} & 1.5 \\
  & Cooperation when opponent cooperated (PD) & prop & \rev{C} & 1.0 \\
  & \rev{Round-to-round RT speed-up} & \rev{corr} & \rev{C} & \rev{1.0} \\
\midrule
\multirow{5}{*}{\texttt{serial\_recall}}
  & Above-chance recall              & prop   & \rev{C} & 1.5 \\
  & Similarity aids recall           & corr   & \rev{C} & 1.5 \\
  & High-vs-low similarity advantage & paired & \rev{C} & 1.0 \\
  & \rev{Similarity RT effect}       & \rev{corr} & \rev{C} & \rev{1.0} \\
  & \rev{Primacy effect}             & \rev{paired} & \rev{C} & \rev{1.0} \\
\midrule
\multirow{3}{*}{\texttt{visual\_recognition}}
  & Above-chance discrimination (hit $>$ FA) & paired & \rev{C} & 2.0 \\
  & Above-chance accuracy            & prop   & \rev{C} & 1.5 \\
  & Low false-alarm rate             & prop   & \rev{C} & 0.5 \\
\midrule
\multirow{4}{*}{\texttt{effort\_foraging}}
  & Travel-cost residence effect (MVT) & \rev{contrast} & \rev{C} & 2.0 \\
  & Above-chance MVT-optimal match   & prop   & \rev{C} & 1.5 \\
  & Above-chance total reward        & prop   & \rev{C} & 1.0 \\
  & \rev{Harvest RT speed-up}        & \rev{corr} & \rev{C} & \rev{1.0} \\
\midrule
\multirow{2}{*}{\texttt{tiny\_alchemy}}
  & Above-chance discovery rate      & prop   & \rev{C} & 2.0 \\
  & Non-redundant attempts           & prop   & \rev{C} & 1.0 \\
\midrule
\rev{\multirow{4}{*}{\texttt{moral\_machine}}}
  & Utilitarian preference           & prop   & \rev{B} & 1.5 \\
  & Save-young preference            & prop   & \rev{B} & 1.0 \\
  & Save-human preference            & prop   & \rev{B} & 1.5 \\
  & Save-lawful preference           & prop   & \rev{B} & 1.0 \\
\midrule
\multirow{4}{*}{\texttt{phishing\_detection}}
  & Above-chance discrimination (hit $>$ FA) & paired & \rev{C} & 2.0 \\
  & Learning effect (post $>$ pre training) & paired & \rev{C} & 1.5 \\
  & Above-chance overall accuracy   & prop   & \rev{C} & 1.0 \\
  & \rev{Practice RT speed-up}      & \rev{corr} & \rev{C} & \rev{1.0} \\
\bottomrule
\end{tabular}
\end{table}

\section{Statistical Test Descriptions}
\label{app:tests}

Level~3 scoring uses four types of statistical tests. Each takes a per-task signature specification (predictor, outcome, expected direction, and weight) and returns a per-signature score in $\{0, 0.5, 1\}$, or reports the signature untestable. Implementations are in \texttt{scoring/analysis\_templates/} in the public repository.

\textbf{One-sample proportion tests} (19 signatures) compare an observed proportion against a chance level with a one-sided exact binomial test, run at the effective sample size $n_{\text{eff}} = n(1-\rho)/(1+\rho)$ given the series' lag-1 autocorrelation $\rho$ (clamped at $n$, so negative autocorrelation cannot add power). They are used for signatures such as above-chance accuracy, Kraken avoidance, and utilitarian preference. The test also computes the smallest $p$ attainable at the given $n$ and chance level. If even a perfect result could not reach $p<0.05$, the signature is reported untestable rather than scored $0$.

\textbf{Paired proportion contrasts} (5 signatures) compare binary accuracy or choice rates between two trial subsets within the same session, taking the $p$-value from a moving-block bootstrap (Appendix~\ref{app:power}) so that within-session dependence does not inflate significance. They are used for the high-vs-low similarity advantage and the primacy effect in serial recall, the post-vs-pre training learning effect in phishing detection, and above-chance discrimination (hit rate vs.\ false-alarm rate) in visual recognition and phishing detection.

\textbf{Conditional proportion contrasts} (2 signatures) compare a binary outcome's rate between two conditions with a one-sided Fisher exact test. The two cases are cooperation after an opponent's cooperation versus after a defection (tit-for-tat reciprocity in repeated games) and the stay rate under high versus low travel cost (the residence effect in effort foraging). A constant policy against a deterministic opponent can leave one condition empty. Reciprocity is then not identifiable from the session, and the signature is reported untestable.

\textbf{Correlation tests} (13 signatures, labeled corr in Table~\ref{tab:signatures}) test the sign and significance of a Pearson correlation between two trial-level variables (e.g., the coherence-accuracy psychometric in random dot motion, EV sensitivity in marbles risk, within-block learning in grid bandit), or a Spearman correlation for the seven response-time signatures, which test relative rather than absolute speed.

Each signature scores 1.0 if the test reaches $p < 0.05$ in the expected direction, 0.5 if it is in the expected direction but non-significant, and 0.0 otherwise. The per-task L3 score is the importance-weighted mean over testable signatures (Equation~\ref{eq:l3}). An undefined statistic is untestable when the predictor is constant, since the design never varied, but scores $0$ when the outcome is constant, since an agent that answers identically at every level of the manipulation shows no effect.

\section{Per-Task Signature Descriptions}
\label{app:signatures}

This appendix describes the signatures evaluated for each of the ten v1 tasks, including the phenomenon each tests and the expected direction from the published literature. Six tasks also carry one response-time relational signature ($w=1.0$, Spearman), in addition to the coherence-RT chronometric signature in \texttt{random\_dot\_motion}, each predicting that response time falls. It should fall across clicks within a block (\texttt{grid\_bandit}), as the expected-value difference grows (\texttt{marbles\_risk}), across rounds (\texttt{repeated\_games}), as cue--target similarity rises (\texttt{serial\_recall}), across harvests within a patch (\texttt{effort\_foraging}), and across trials within a phase (\texttt{phishing\_detection}). These signatures test relative speed within a session, which survives the uniform slowing that the model's response time imposes (Section~\ref{sec:discussion}). The paragraphs below describe the remaining signatures in Table~\ref{tab:signatures}.

\paragraph{\texttt{random\_dot\_motion}} (after \citealp{pirrone2018evidence, roitman2002response}). The coherence-accuracy psychometric tests whether choice accuracy increases monotonically with motion coherence ($w=2.0$). The coherence-RT chronometric tests whether correct response times decrease as coherence increases ($w=1.5$). Above-chance accuracy at the lowest non-zero coherence ($w=0.5$) and overall above-chance accuracy ($w=1.0$) test whether the motion-direction signal is extracted at all.

\paragraph{\texttt{grid\_bandit}} (after \citealp{wu2018generalization, schulz2019searching}). Kraken avoidance in risky blocks ($w=1.5$) tests whether the agent skirts tiles whose underlying value exceeds a hidden ``Kraken'' threshold. Value-directed exploitation ($w=1.5$) tests whether high-reward clicks attract spatially nearby follow-up clicks. Above-chance reward ($w=1.0$) tests whether mean reward exceeds the grid mean of 50, and within-block learning ($w=1.0$) tests whether observed reward grows monotonically across click index within a block.

\paragraph{\texttt{marbles\_risk}} (after \citealp{levin2003risk, tymula2013like}). EV sensitivity ($w=2.0$) tests whether the probability of choosing the risky urn correlates with the EV difference between the risky and safe options. Above-chance EV-optimal choice ($w=1.5$) tests whether the agent picks the higher-EV option at rates above chance on trials with a clear EV advantage. The magnitude effect ($w=1.0$) tests whether the risky option is chosen more often when its payoff is larger, across matched probability levels.

\paragraph{\texttt{repeated\_games}} (after \citealp{akata2025playing} for the repeated PD against tit-for-tat and \citealp{cooper1989communication} for BoS coordination). Tit-for-tat reciprocity ($w=1.5$) tests whether the agent's current cooperate decision tracks the opponent's previous-round move in the PD blocks. Above-chance coordination ($w=1.5$) tests whether the agent matches the opponent's BoS action at rates above chance. Conditional cooperation when the opponent cooperated ($w=1.0$) tests whether, when the opponent cooperated in the previous round, the agent's cooperation rate exceeds 0.5 (the v1.1 audit replaced an earlier non-zero-cooperation signature that random play trivially passed).

\paragraph{\texttt{serial\_recall}} (after \citealp{murdock1962serial, haridi2026context}). Above-chance recall ($w=1.5$) tests whether overall recall accuracy exceeds the typed-three-letters chance baseline. Similarity-aids-recall ($w=1.5$) tests whether recall accuracy increases with semantic similarity of the cue-target pair. The high-vs-low similarity advantage ($w=1.0$) sharpens this contrast as a paired proportion test on the extreme similarity buckets. The primacy effect ($w=1.0$) tests whether pairs studied in the first quarter of the list are recalled better than those studied in the middle. Position bands are defined from the configured list length of 16 pairs rather than the highest position a session reached, so a truncated session cannot relabel positions. Recency is not scored, for the reason given in Appendix~\ref{app:rep_signatures}.

\paragraph{\texttt{visual\_recognition}} (after \citealp{brady2008visual, standing1973learning}). Above-chance discrimination ($w=2.0$) tests whether the hit rate on old items significantly exceeds the false-alarm rate on new items, the canonical recognition-memory comparison. Above-chance accuracy ($w=1.5$) tests overall accuracy against chance. Low false-alarm rate ($w=0.5$) tests whether the agent avoids the trivial old-bias strategy.

\paragraph{\texttt{effort\_foraging}} (after \citealp{constantino2015learning, bustamante2023effort, charnov1976optimal}). The marginal value theorem (MVT) says that a forager should leave a patch when its current rate of reward falls below the average rate for the environment. The travel-cost residence effect ($w=2.0$) tests whether the agent stays in a patch more often under high than under low travel cost between patches, the central MVT prediction. It is scored from the stay rate on every trial rather than from residence times, because agents leave patches too rarely for residence times to be testable. Above-chance MVT match ($w=1.5$) tests whether per-trial stay/leave decisions match the MVT-optimal prediction. Above-chance total reward ($w=1.0$) tests whether reward per decision exceeds the random-policy baseline.

\paragraph{\texttt{tiny\_alchemy}} (after \citealp{brandle2023empowerment}). Above-chance discovery ($w=2.0$) tests whether successful element combinations occur at a rate above the empirical random-agent ceiling. Non-redundant attempts ($w=1.0$) tests whether the proportion of unique unordered \texttt{(element\_a, element\_b)} pairs across the session exceeds the random-policy ceiling. The v1.1 audit dropped an empowerment-preference correlation that was firing mechanically for random play (high in-degree base elements were over-sampled by chance) without indicating cognitive selectivity.

\paragraph{\texttt{moral\_machine}} (after \citealp{awad2018moral}). Utilitarian preference ($w=1.5$) tests whether agents prefer to save the larger group when group sizes differ. Save-young ($w=1.0$) tests the cross-cultural preference for saving the young over the elderly. Save-human ($w=1.5$) tests the preference for saving humans over pets, the largest single-factor effect in the original Moral Machine corpus. Save-lawful ($w=1.0$) tests the preference to save lawful over unlawful crossers ($0.68$ in Awad et al.'s data). \rev{An intervention-aversion signature ($w=0.5$), testing the mild human preference against swerving when outcomes are otherwise equivalent ($0.47$ intervention rate in Awad et al.'s data), was removed from the v1 specification after the power audit found it untestable at $n=4$ (Appendix~\ref{app:power}). No metric scores it now, since \texttt{moral\_machine} has no Level~2 metrics (Appendix~\ref{app:metrics}).}

\paragraph{\texttt{phishing\_detection}} (after \citealp{singh2019training}). Above-chance discrimination ($w=2.0$) tests whether the hit rate on phishing emails exceeds the false-alarm rate on legitimate emails. The learning effect ($w=1.5$) tests whether accuracy on the ten post-training trials significantly exceeds accuracy on the ten pre-training trials. Above-chance overall accuracy ($w=1.0$) tests whether classification accuracy across all sixty trials is above chance.

\section{Level 2 Accuracy Metrics in v1}
\label{app:metrics}

CogArena v1 evaluates \rev{31} Level~2 accuracy metrics across \rev{eight} of the ten tasks, \rev{3.9} per task on average. \rev{Absolute response time is not among them. It is scored at Level~3 through relational signatures instead (Section~\ref{sec:discussion}).} \rev{The normalization is monotonic in $z$, which rewards exceeding the human value. That is correct only for a metric on which exceeding the human value is better performance, so Level~2 is restricted to such metrics. Three groups that the submitted version scored here are therefore not scored at Level~2. The five \texttt{moral\_machine} proportions (utilitarian, save-young, save-human, save-lawful, and intervention choices) are preferences with no better direction, so the monotonic mapping rewarded an agent for being more extreme than people, and each preference was also counted a second time through its Level~3 signature. The rate of higher-expected-value choices in \texttt{marbles\_risk} treats risk neutrality as correct, which expected-utility theory does not, since it leaves the decision-maker's risk attitude open. Similarly, the stay rate and the two mean residence times in \texttt{effort\_foraging} rewarded staying longer, but under the marginal value theorem staying is better only up to the optimum, so these metrics rewarded over-harvesting. Agents stayed on $96\%$ of decisions and scored higher on the stay rate than on any other foraging metric. Whether an agent shares the human direction on each of these patterns is tested at Level~3. As a result, \texttt{moral\_machine} and \texttt{marbles\_risk} have no Level~2 score, and their composite redistributes the Level~2 weight to Levels~1 and~3 in proportion (Section~\ref{sec:scoring}). Five metrics are scored with the sign reversed, because for them a lower value is better performance. These are the two false-alarm rates, \texttt{grid\_bandit}'s Kraken-caught rate, and the two timeout rates.} Each metric is normalized against its source-study human baseline (Appendix~\ref{app:provenance}) by converting it to a $z$-score and mapping the $z$-score through the standard normal CDF to a score between 0 and 1 (Section~\ref{sec:scoring}). The per-task metrics are listed below, and full specifications, including human baseline means and standard deviations, are in \texttt{tasks/<task\_id>/scoring/level2\_metrics.json} in the public repository.

\begin{itemize}
  \item \texttt{random\_dot\_motion} (\rev{4} metrics) covers overall accuracy and accuracy at low, medium, and high coherence.
  \item \texttt{grid\_bandit} (5 metrics) covers mean reward per click, mean reward in safe and in risky blocks, the proportion of high-value clicks\rev{ (tested against an explicit numeric threshold)}, and the Kraken-caught rate in risky blocks.
  \item \texttt{marbles\_risk} has \rev{no Level~2 metrics. Its higher-EV choice rate is tested at Level~3.}
  \item \texttt{repeated\_games} (5 metrics) covers the cooperation rate in the PD, the coordination rate in the BoS, mean payoff in each game, and the timeout rate.
  \item \texttt{serial\_recall} (\rev{4} metrics) covers overall accuracy, accuracy at high and at low similarity, and the timeout rate.
  \item \texttt{visual\_recognition} (\rev{3} metrics) covers overall accuracy, hit rate, and false-alarm rate.
  \item \texttt{effort\_foraging} (\rev{2} metrics) covers total reward and the MVT-optimal match rate.
  \item \texttt{tiny\_alchemy} (3 metrics) covers success rate, novelty rate, and unique-pair rate.
  \item \texttt{moral\_machine} has \rev{no Level~2 metrics. Its preferences are tested at Level~3.}
  \item \texttt{phishing\_detection} (5 metrics) covers overall accuracy, pre- and post-training accuracy, hit rate, and false-alarm rate.
\end{itemize}

\section{\rev{Human-Baseline Provenance}}
\label{app:provenance}

\rev{Every Level~3 expected direction in v1, and every Level~2 baseline except four targets that we set ourselves, comes from published work rather than from participants completing our versions of the tasks. Because a measured agent--human difference could in principle reflect our implementation rather than the agent, this appendix documents, for each task, the source of the baseline, how firmly it is anchored in that source, and the specific ways our implementation differs from the study the numbers come from. Twenty-four of the thirty-one Level~2 baselines, spanning six tasks, are computed directly from the original authors' released participant data. Of the other seven, three are \texttt{visual\_recognition} baselines derived from the source paper's reported statistics, because no public archive of that study exists. Two are \texttt{effort\_foraging} targets set from the design of our own version of the task, because the source's travel manipulation differs from ours. The last two are timeout-rate targets, which we set ($0.02$ for \texttt{repeated\_games} and $0.05$ for \texttt{serial\_recall}) rather than measured from participants. Because deadlines were disabled for agent sessions (Appendix~\ref{app:harness}), no agent trial could time out, and these two metrics took the same value in every agent session. The \texttt{marbles\_risk} and \texttt{moral\_machine} rates and the \texttt{effort\_foraging} stay rate are also computed from released data, but they are preferences or strategy descriptors rather than accuracy measures, so they are kept as reference values and not scored (Appendix~\ref{app:metrics}). The values actually used in scoring are the \texttt{human\_mean} and \texttt{human\_sd} fields of each task's \texttt{tasks/<task>/scoring/level2\_metrics.json}. The source archives ship with the release, together with per-file citations, and \texttt{scripts/derive\_baselines.py} documents, for every derived value, the archive, condition, and participants it comes from.}

\rev{The Tier column of Table~\ref{tab:provenance} records how each baseline is derived. R (raw data) means the value is computed directly from the original authors' released participant data, and six tasks carry this tier throughout. D (\texttt{effort\_foraging}) means the targets are set from the design of our version of the task. A dash marks a task with no Level~2 metrics, whose source supplies reference values only. P (paper-derived) means no public archive exists and the value is derived from statistics reported in the source paper, which applies only to \texttt{visual\_recognition}. Earlier versions used author estimates throughout; the current release replaces them with recomputed values wherever source data exist, and \texttt{scripts/derive\_baselines.py} reproduces every derived value from the archived data.}

\begin{table}[h]
\caption{\rev{Provenance of the v1 human baselines. In the Tier column, R means computed from the original authors' released participant data, D means set from the design of our version of the task, P means derived from statistics published in the source paper because no public archive is available, and a dash marks a task with no Level~2 metrics.}}
\label{tab:provenance}
\centering
\revcolor
\footnotesize
\resizebox{\textwidth}{!}{%
\begin{tabular}{@{}llcp{7.6cm}@{}}
\toprule
\textbf{Task} & \textbf{Baseline source} & \textbf{Tier} & \textbf{Port differences relevant to the baseline} \\
\midrule
\texttt{random\_dot\_motion} & \rev{\citealt{desender2021dynamic} (Confidence Database) and \citealt{roitman2002response} (paradigm)} & \rev{R} & \rev{Accuracy baselines computed from trial-level human data at coherence levels matched to ours. Our Canvas implementation reproduces the Roitman and Shadlen algorithm but not their apparatus or viewing distance.} \\
\texttt{grid\_bandit} & \rev{\citealt{witte2024exploring}} & \rev{R} & \rev{Computed from the authors' participant data (26{,}565 clicks, $n=220$), replacing the GP-bandit literature expectations used at submission.} \\
\texttt{marbles\_risk} & \rev{\citealt{ciranka2022uncertainty}} & \rev{--} & \rev{Higher-EV choice rate computed from the authors' trial-level data (23{,}887 trials, $n=166$), restricted to the description condition our version implements. A reference value only, since the task has no Level~2 metrics.} \\
\texttt{repeated\_games} & \rev{\citealt{akata2025playing}} & \rev{R} & \rev{Cooperation and coordination rates computed from the source's human rounds (3{,}900 rounds, $n=195$) against the same simple opponents our port uses.} \\
\texttt{serial\_recall} & \rev{\citealt{haridi2026context}} & \rev{R} & \rev{Computed from the authors' trial-level data for the 16-pair lists our version uses ($n=29$ of the archive's $116$ participants). The source used $\sim$28 pairs per similarity level and a typed, spacebar-gated response, whereas our version uses 16 pairs across four similarity levels with a single typed recall step.} \\
\texttt{visual\_recognition} & \citealt{brady2008visual} & \rev{P} & \rev{No public archive exists. The source's memory test is two-alternative forced choice (92.5\%, 87.6\%, and 87.2\% across novel, exemplar, and state conditions). Old/new hit and false-alarm baselines are derived from those statistics through $d'$, assuming an unbiased criterion, so that hit rate $= \Phi(d'/2)$ and false-alarm rate $= 1 - \Phi(d'/2)$; the derivation is recorded in \texttt{scoring/human\_baselines/visual\_recognition.json}. Our version uses 768 procedurally generated SVG stimuli (shape $\times$ color $\times$ pattern $\times$ orientation), which are plausibly easier to discriminate, so this baseline is likely too lenient.} \\
\texttt{effort\_foraging} & \rev{\citealt{constantino2015learning} (design) and \citealt{bustamante2023effort} (data)} & \rev{D} & \rev{Reward and MVT-match targets set from the design of our version, since the source's travel manipulation differs from ours. The stay rate computed from the authors' trial-level data (350{,}608 decisions, $n=537$) is a reference value only.} \\
\texttt{tiny\_alchemy} & \citealt{brandle2023empowerment} & \rev{R} & \rev{Computed from the authors' trial data for the Tiny Alchemy experiment itself (48{,}963 trials, $n=97$), the variant our version implements, rather than the larger Little Alchemy 2 dataset. Our port caps attempts at 200 or twelve minutes.} \\
\texttt{moral\_machine} & \rev{\citealt{awad2018moral}} & \rev{--} & \rev{Preference rates computed from the authors' published effect-size archive ($3.52\times10^{7}$ decisions), re-expressed as proportions in $[0,1]$ rather than the paper's centered effect sizes. Reference values only, since the task has no Level~2 metrics.} \\
\texttt{phishing\_detection} & \rev{\citealt{singh2019training}} & \rev{R} & \rev{Computed from the authors' trial-level data restricted to the outcome-feedback, 50\% base-rate condition that matches our design ($n=99$ of the archive's $296$ participants).} \\
\bottomrule
\end{tabular}%
}
\end{table}

\rev{Two consequences follow for interpretation. First, the two levels are unequally exposed to this provenance gap. Level~3 tests the direction of an effect, which transfers across implementations far better than magnitude does. Level~2 converts a raw value into a $z$-score against a specific mean and standard deviation, so an implementation that shifts the underlying distribution shifts the score even when behavior is unchanged. Level~2 should accordingly be read as normalization against a source-study anchor, not as a comparison against humans on this platform. Second, the direction of the likely distortion is known in at least one case. Our \texttt{visual\_recognition} stimuli are compositionally structured and therefore probably easier than the photographs in the source study, so a human completing our version would likely exceed the baseline we score against, which makes agent scores on that task optimistic rather than pessimistic. Collecting on-platform human data for all ten tasks is the planned remedy and would replace every row of Table~\ref{tab:provenance}.}

\section{\rev{Run Settings and What the Model Saw}}
\label{app:harness}

\rev{The platform does not constrain how an agent drives the browser (Section~\ref{sec:methods}), so the settings below describe our study rather than the benchmark. We record them because a score belongs to a model together with its scaffold, and because what the model is shown determines which task information it can use at all.}

\rev{\paragraph{Scaffold.} All four evaluated models ran under Browser-Use~\citep{browseruse2024}, with model calls routed through OpenRouter. Screenshots were enabled, and each agent was limited to $5$ actions per step, $10$ consecutive failed actions before the run was stopped, $500$ steps, and $1800$\,s per task, with the same limits for every task. Browser-Use's built-in check for an agent repeating the same action was left on, and we logged any second visit to a task's web page, since reloading the page restarts the jsPsych experiment and discards the trials recorded so far.}

\rev{\paragraph{What the model saw.} With screenshots enabled, Browser-Use gives the model at each step a screenshot of the visible part of the page together with a text list of the page's clickable and typeable elements. Every step of every evaluated session used this combination; no model was ever reduced to the text list alone. The Random Agent used for the chance floor has no language model and reads only the page text, through the Playwright browser-automation library. That is enough to press a random valid key on each trial, but it means the floor was not collected under the same viewing conditions as the evaluated agents.}

\rev{\paragraph{Response deadlines.} Agent sessions were run with per-trial response deadlines switched off (the \texttt{no\_deadline} option in the task's web address), so that a slow model response could not by itself count as a missed trial. jsPsych records the actual response time regardless of the deadline setting, so Level~3 response-time signatures remain well defined. Absolute response time is not scored at Level~2, for the reasons given in Section~\ref{sec:discussion}.}

\begin{figure}[h]
\centering
\includegraphics[width=\textwidth]{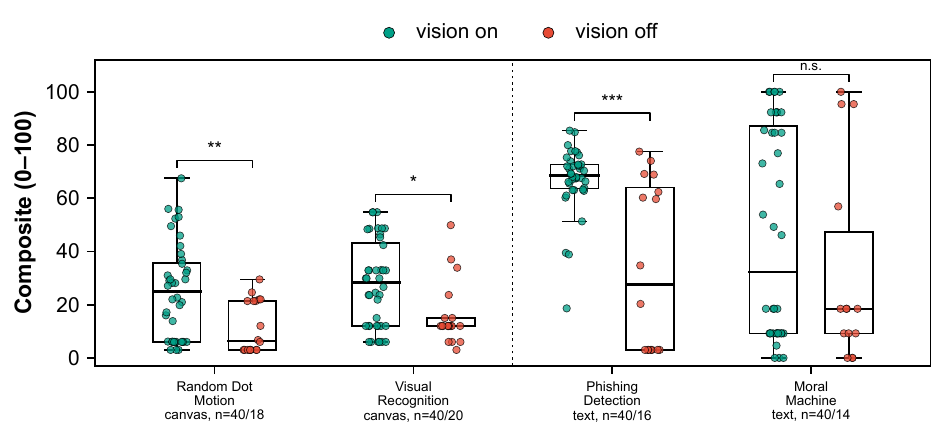}
\caption{\rev{Screenshot-removal check. Same models and scaffold, with screenshots removed so the model received only the page text. Each boxplot shows the per-task composite under one condition, with one point per scored session. The $n$ is given as with/without screenshots. Tasks are grouped by whether their stimuli are drawn as images, which the page text does not describe, or are themselves text. Brackets give two-sided Mann-Whitney tests between conditions ($^{*}p<0.05$, $^{**}p<0.01$, $^{***}p<0.001$, n.s.\ not significant). All four tasks degrade, three of them significantly. The two text tasks start two to three times higher than the two image tasks, so the raw drops shown here are not comparable across the two groups, but the proportional changes quoted in Section~\ref{sec:design} are.}}
\label{fig:ablation}
\end{figure}

\rev{\paragraph{Screenshot-removal check.} To test whether the agents actually use what they see, rather than assume it, we re-ran the same models under the same scaffold and scoring with the screenshots removed, so that the model received only the page text. We chose four tasks for a deliberate contrast. The \texttt{random\_dot\_motion} and \texttt{visual\_recognition} tasks draw their stimuli as images, which the page text does not describe, while \texttt{phishing\_detection} and \texttt{moral\_machine} present their content as text. If screenshots mattered mainly for reading the stimuli, the first pair should collapse and the second should not.}

\rev{Removing the screenshots degraded all four tasks, by $31$ to $53\%$ of composite, and significantly in three (Figure~\ref{fig:ablation}). The largest single drop was \texttt{random\_dot\_motion} at $-53\%$, the task whose stimulus is least recoverable from the page text, which is what one would expect if screenshots are used to read the stimuli. The rest of the pattern is not. The \texttt{visual\_recognition} task fell by $43\%$ while \texttt{phishing\_detection}, whose content is fully present as text, fell by $49\%$, and the gap between the image and text tasks ($-48\%$ against $-40\%$) rests on the one task that did not degrade significantly. The result therefore supports the claim that agents on CogArena depend on the screenshots, but not the stronger claim that they use them to read the stimuli. The simpler reading is that these agents use screenshots for general interface orientation and lose that support wherever it is removed. Proportional rather than raw change is the appropriate quantity here, because the text tasks start two to three times higher and have more room to fall.}

\rev{\paragraph{Temporal stimuli.} One v1 task carries its manipulated variable across frames rather than within one. The \texttt{random\_dot\_motion} task draws $200$ moving dots for $2000$\,ms per trial, with the coherent fraction translating at $3$\,px per frame in the signaled direction while the remainder move in random directions. Motion coherence is thus a property of the sequence of frames and is absent from any single frame. An agent that sees one screenshot per step, with steps far longer than a frame, receives no information about the coherence manipulation, and its scores on this task should be interpreted accordingly. Recovering such stimuli requires a scaffold that captures video or several screenshots per second, which the platform permits but our study did not use.}

\section{\rev{Signature Trial Counts and Statistical Power}}
\label{app:power}

\rev{Level~3 scores each signature from a single session's trials, so the number of trials feeding a given test bounds what that test can detect. Table~\ref{tab:trialcounts} gives the trial budget per task and the subset each signature draws on. The spread is wide. The coherence psychometric in \texttt{random\_dot\_motion} draws on $100$ trials, whereas several signatures in \texttt{moral\_machine} and \texttt{phishing\_detection} draw on a few dozen or fewer, because the trial budget is divided across the dimensions or blocks the signature contrasts.}

\begin{table}[h]
\caption{\rev{Trial budgets available to Level~3 signatures in v1. ``Per-signature $n$'' is the number of trials the corresponding tests draw on after filtering.}}
\label{tab:trialcounts}
\centering
\revcolor
\small
\begin{tabular}{@{}lrp{7.2cm}@{}}
\toprule
\textbf{Task} & \textbf{Trials} & \textbf{Per-signature $n$} \\
\midrule
\texttt{random\_dot\_motion} & 100 & 100 overall, 20 per coherence level (5 levels, 5 blocks); the configuration file lists 120, but the implementation balances 24 trials per block over ten coherence--direction cells and so presents 20, and Level~1 was scored against the configured 120 \\
\texttt{grid\_bandit}        & 110 & 11 blocks $\times$ 10 clicks, with 50 risky-block clicks for Kraken avoidance \\
\texttt{marbles\_risk}       & 64  & 64 overall, 4 repetitions of the gamble grid \\
\texttt{repeated\_games}     & 30  & 15 PD rounds and 15 BoS rounds, and conditional signatures draw on the subset of PD rounds following a given opponent move \\
\texttt{serial\_recall}      & 16  & 16 recall trials across 4 similarity levels \\
\texttt{visual\_recognition} & 100 & 100 test trials (50 old, 50 lures) after 50 study trials \\
\texttt{effort\_foraging}    & 80  & 2 blocks $\times$ 40 trials, 40 per travel-cost condition \\
\texttt{tiny\_alchemy}       & $\le$200 & up to 200 combination attempts or 12 minutes, from 4 base elements (the random agent runs 80, so its floor on this task is a reference at a smaller budget rather than a matched comparison) \\
\texttt{moral\_machine}      & 40  & 10 each for number, age, and species, 6 for legality, and 4 intervention trials that no signature uses \\
\texttt{phishing\_detection} & 60  & 60 overall, 10 pre-training and 10 post-training for the learning effect \\
\bottomrule
\end{tabular}
\end{table}

\rev{This appendix examines the power of the signature tests under short trial budgets, the sensitivity of the results to within-session dependence, and how many signature observations could not be tested.}

\rev{\paragraph{Power.} The shortest signatures cannot detect the effects they target. Detecting a preference of the size in Awad et al.'s data for legality ($0.68$ against a chance level of 0.5) at 80\% power needs about fifty trials on that dimension, and v1 allocates six. A null on such a signature is therefore weak evidence that the agent lacks the pattern. Intervention aversion in \texttt{moral\_machine} was worse than underpowered. It ran on four trials, where no result can reach $p<0.05$, so it was untestable in every session and has been removed from the specification.}

\rev{\paragraph{Dependence.} The tests assume independent trials, and agents violate that assumption more than human participants do. A Browser-Use agent carries its own action history in context and readily settles into a repeated policy. Across the archive the median lag-1 autocorrelation is near zero, but the tail is severe. The 90th percentile is $\rho=0.75$, the maximum is $0.93$, and $23.5\%$ of proportion tests exceed $\rho=0.3$, against $1.6\%$ for the random agent. We therefore checked Level~3 against a moving-block bootstrap, which imposes no parametric model of the dependence. Contiguous blocks of trials are resampled, so dependence within a block is preserved while dependence across blocks is broken, and the $p$-value inverts the resulting percentile interval. Re-running the $920$ signature tests in measurable sessions this way changed $5.5\%$ of grades, left the model ranking unchanged, and slightly widened rather than narrowed the gap between the agents and the chance floor ($+0.013$ on Level~3). Fixed ten-trial blocks, well above the default length of two to five, changed more grades ($12.2\%$) and widened the gap further ($+0.018$). The all-session composite ranking was unchanged under both block rules. Over measurable sessions only, ten-trial blocks moved \texttt{qwen3-vl-235b-instruct} above \texttt{gemini-3-flash-preview} on Level~3 alone; \texttt{gemini-3-flash-preview} still leads on the headline composite under both conventions (Section~\ref{sec:study}). The comparison also shows that the first-order correction $n_{\text{eff}} = n(1-\rho)/(1+\rho)$ applied to the one-sample proportion tests over-charges for dependence, since on that test $26$ grades rise under the bootstrap and none fall. It showed the opposite for the paired proportion contrast, whose original pooled normal approximation was anti-conservative (median $p$ $2.4\times$ too small). The reported scores therefore take that test's $p$-value from the bootstrap itself, and the switch changed four grades across the archive. Neither treatment separates serial dependence from a genuine within-session trend, so a learning curve is penalized like a sticky response. On the whole, then, these corrections are more likely to lower Level~3 than to raise it, although this need not hold for every signature.}

\rev{\paragraph{Untestable observations.} Of the $1{,}560$ signature observations in the panel, $497$ ($32\%$) were untestable, mostly because a session ended early or the agent answered identically throughout. Table~\ref{tab:sig_pass} gives the testable count for every signature. Averaged over all four hundred sessions, with a Level~3 of zero for the $83$ sessions in which no signature could be tested, the mean Level~3 is $0.365$, against $0.490$ over measurable sessions.}

\rev{\paragraph{Consequence for reading Level~3.} Because each cell is ten sessions and the discrete $\{0,0.5,1\}$ scoring moves a full step at the significance boundary, per-cell values remain noisy even after these corrections (Section~\ref{sec:study_gap}). Increasing trial budgets on the short signatures is the change that would most improve Level~3's resolution, and it requires a task revision rather than a scoring one.}

\section{Chance-Floor Analysis}
\label{app:chance}

We ran the Random Agent (a script with no language model that presses a random valid key on every trial) ten times per task, \rev{and all one hundred sessions were scored}. Between the first chance-floor run (v1.0) and the one reported here, a v1.1 audit tightened two signatures and dropped a third in \texttt{tiny\_alchemy}, and replaced one in \texttt{repeated\_games}, all of which random responding had passed too easily (Appendix~\ref{app:signatures}). Their revised specifications ship with the tagged release, and the changes are recorded in the repository changelog. Table~\ref{tab:floor} reports per-task chance-floor scores. \rev{The floor is scored by the same code as the agents, so the two are directly comparable. Across all one hundred sessions the mean Level~3 is $0.184$ and the mean composite is $31.31$.}

\begin{table}[h]
\caption{Random-agent chance floor for each v1 task, \rev{mean over ten sessions per task}. All L1 values are 1.0 because the driver always reaches end-of-experiment. \rev{A dash marks a task with no Level~2 metrics, whose composite redistributes the Level~2 weight to Levels~1 and~3. That raises those two floors, because the random agent always completes.}}
\label{tab:floor}
\centering
\small
\begin{tabular}{@{}lccccc@{}}
\toprule
\textbf{Task} & \textbf{$n$} & \textbf{L1} & \textbf{L2} & \textbf{L3} & \textbf{Composite} \\
\midrule
\texttt{random\_dot\_motion}  & \rev{10} & 1.000 & \rev{0.024} & \rev{0.335} & \rev{32.60} \\
\texttt{grid\_bandit}             & \rev{10} & 1.000 & \rev{0.045} & \rev{0.179} & \rev{25.54} \\
\texttt{marbles\_risk}            & \rev{10} & 1.000 & \rev{--} & \rev{0.191} & \rev{37.76} \\
\texttt{repeated\_games}          & \rev{10} & 1.000 & \rev{0.428} & \rev{0.217} & \rev{40.86} \\
\texttt{serial\_recall}       & \rev{10} & 1.000 & \rev{0.183} & \rev{0.050} & \rev{23.92} \\
\texttt{visual\_recognition}      & \rev{10} & 1.000 & \rev{0.001} & \rev{0.331} & \rev{31.59} \\
\texttt{effort\_foraging}         & \rev{10} & 1.000 & \rev{0.237} & \rev{0.186} & \rev{32.60} \\
\texttt{tiny\_alchemy}            & \rev{10} & 1.000 & \rev{0.298} & 0.000 & \rev{25.42} \\
\texttt{moral\_machine}           & \rev{10} & 1.000 & \rev{--} & \rev{0.191} & \rev{37.79} \\
\texttt{phishing\_detection}  & \rev{10} & 1.000 & \rev{0.059} & \rev{0.159} & \rev{25.03} \\
\midrule
\textbf{Mean across v1}            & -- & 1.000 & \rev{0.159} & \rev{0.184} & \rev{31.31} \\
\bottomrule
\end{tabular}
\end{table}

\rev{Two properties of the floor matter for interpretation. Its Level~2 is low ($0.159$, over the eight tasks that have one) because a random agent has no way to be accurate, and because the metrics it could once score well on by responding instantly are no longer part of Level~2. Its Level~3 of $0.184$ is not zero, since a handful of directional tests will reach significance on noise across a hundred sessions, which is exactly the quantity a floor exists to make visible. The floor also certifies the benchmark's discriminative property, since no signature was passed by the random agent in at least half its sessions.}

\section{Per-Task Results}
\label{app:per_task}

\begin{figure}[h]
\centering
\includegraphics[width=\textwidth]{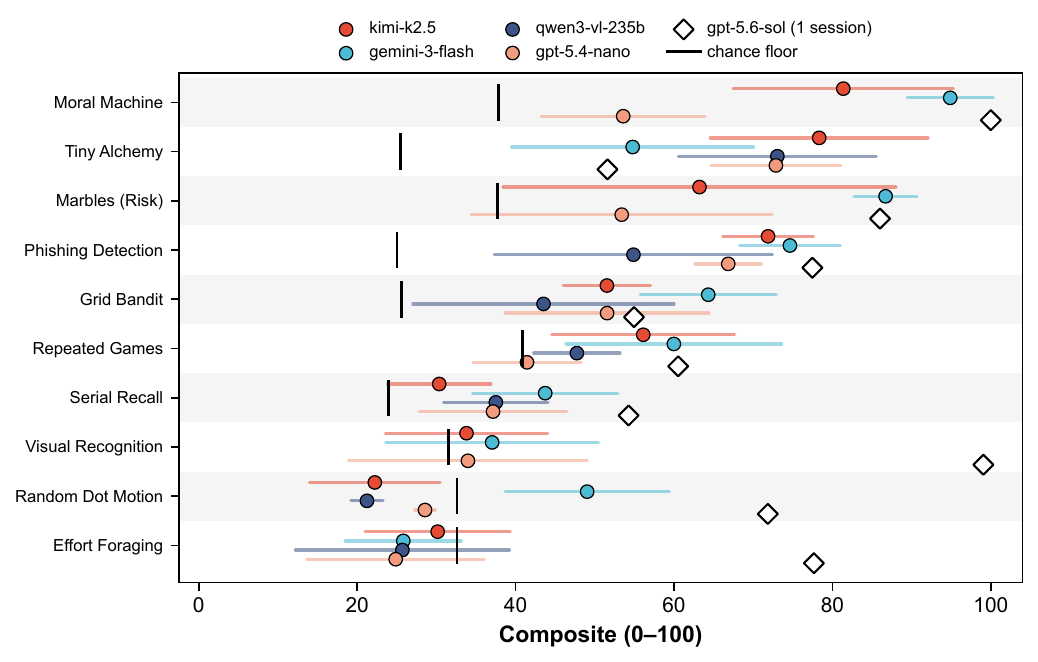}
\caption{Per-task composite. Colored points are panel models, mean $\pm$ SD over each cell's measurable sessions (at most ten). Each occupies a fixed vertical slot within its task band, so identity is carried by position as well as color. \rev{The open diamond is the state-of-the-art model \texttt{gpt-5.6-sol}, drawn without an interval because it is a single session and shown as a reference rather than a fifth panel member (Appendix~\ref{app:flagship}).} The vertical rule is the re-scored per-task chance floor. Tasks are ordered by panel mean. Overlapping intervals on most tasks are the reason per-cell values do not support model comparisons at this sample size.}
\label{fig:taskprofile}
\end{figure}

Table~\ref{tab:per_task_full} reports per-task composite scores for every model in the panel alongside the random-agent floor, as the mean over that cell's measurable sessions (at most ten) with the standard deviation across them. Figure~\ref{fig:taskprofile} plots the same values.

Two patterns are visible here that the aggregate distributions of Figure~\ref{fig:scores} cannot show. First, tasks differ widely in difficulty. Every model clears the floor comfortably on \texttt{phishing\_detection}, by \rev{$30$ to $50$} points, while on \texttt{visual\_recognition} the best model sits \rev{$5.4$} points above it \rev{and the weakest model with measurable sessions $2.2$}. Second, the ranking is not uniform across tasks. The \texttt{gemini-3-flash-preview} model leads the panel overall but is beaten by \rev{every other panel member} on \texttt{tiny\_alchemy}, and its margin on \texttt{marbles\_risk} (\rev{$+23.5$} over the runner-up) is larger than its whole-panel margin. \rev{On \texttt{effort\_foraging} every model scores below the random floor. Agents almost never leave a patch, so they match the MVT-optimal decision less often than random choice does and earn less reward.}

The dispersions are the more important column. Within-cell standard deviations reach \rev{$24.8$} on \texttt{marbles\_risk}, and intervals for different models overlap on most tasks even where their means differ by twenty points or more. Per-task, per-model comparisons at this sample size are therefore not identified, which is why Section~\ref{sec:study_gap} restricts its claims to aggregates and why we report dispersion rather than means alone.

\begin{table}[h]
\caption{\rev{Per-task composite over measurable sessions, mean $\pm$ SD, against the random-agent floor. A dash marks a cell with no measurable session. Rows follow the task order of Figure~\ref{fig:taskprofile}. The final rows give each model's mean over its measurable sessions and the number of such sessions, which differs sharply across models and is the reason per-task means alone do not rank them.}}
\label{tab:per_task_full}
\centering
\small
\begin{tabular}{@{}lrrrrr@{}}
\toprule
\textbf{Task} & \textbf{Floor} & \textbf{\texttt{gemini-3-flash}} & \textbf{\texttt{kimi-k2.5}} & \textbf{\texttt{gpt-5.4-nano}} & \textbf{\texttt{qwen3-vl-235b}} \\
\midrule
\texttt{moral\_machine} & \rev{37.8} & \rev{94.9 {\scriptsize $\pm$ 5.4}} & \rev{81.4 {\scriptsize $\pm$ 13.9}} & \rev{53.6 {\scriptsize $\pm$ 10.3}} & -- \\
\texttt{tiny\_alchemy} & \rev{25.4} & \rev{54.8 {\scriptsize $\pm$ 15.3}} & \rev{78.3 {\scriptsize $\pm$ 13.7}} & \rev{72.9 {\scriptsize $\pm$ 8.2}} & \rev{73.0 {\scriptsize $\pm$ 12.5}} \\
\texttt{marbles\_risk} & \rev{37.8} & \rev{86.7 {\scriptsize $\pm$ 4.0}} & \rev{63.2 {\scriptsize $\pm$ 24.8}} & \rev{53.4 {\scriptsize $\pm$ 19.0}} & -- \\
\texttt{phishing\_detection} & \rev{25.0} & \rev{74.6 {\scriptsize $\pm$ 6.4}} & \rev{71.9 {\scriptsize $\pm$ 5.7}} & \rev{66.8 {\scriptsize $\pm$ 4.2}} & \rev{54.9 {\scriptsize $\pm$ 17.6}} \\
\texttt{grid\_bandit} & \rev{25.5} & \rev{64.3 {\scriptsize $\pm$ 8.6}} & \rev{51.5 {\scriptsize $\pm$ 5.5}} & \rev{51.6 {\scriptsize $\pm$ 12.8}} & \rev{43.5 {\scriptsize $\pm$ 16.5}} \\
\texttt{repeated\_games} & \rev{40.9} & \rev{60.0 {\scriptsize $\pm$ 13.6}} & \rev{56.1 {\scriptsize $\pm$ 11.5}} & \rev{41.4 {\scriptsize $\pm$ 6.8}} & \rev{47.7 {\scriptsize $\pm$ 5.4}} \\
\texttt{serial\_recall} & \rev{23.9} & \rev{43.7 {\scriptsize $\pm$ 9.2}} & \rev{30.4 {\scriptsize $\pm$ 6.5}} & \rev{37.2 {\scriptsize $\pm$ 9.3}} & \rev{37.5 {\scriptsize $\pm$ 6.6}} \\
\texttt{visual\_recognition} & \rev{31.6} & \rev{37.0 {\scriptsize $\pm$ 13.4}} & \rev{33.8 {\scriptsize $\pm$ 10.2}} & \rev{34.0 {\scriptsize $\pm$ 15.1}} & -- \\
\texttt{random\_dot\_motion} & \rev{32.6} & \rev{49.0 {\scriptsize $\pm$ 10.4}} & \rev{22.2 {\scriptsize $\pm$ 8.2}} & \rev{28.6 {\scriptsize $\pm$ 1.3}} & \rev{21.2 {\scriptsize $\pm$ 2.0}} \\
\texttt{effort\_foraging} & \rev{32.6} & \rev{25.8 {\scriptsize $\pm$ 7.3}} & \rev{30.2 {\scriptsize $\pm$ 9.2}} & \rev{24.9 {\scriptsize $\pm$ 11.2}} & \rev{25.7 {\scriptsize $\pm$ 13.5}} \\
\midrule
\textbf{Mean (measurable)} & \rev{31.3} & \rev{\textbf{58.7}} & \rev{\textbf{50.7}} & \rev{\textbf{45.7}} & \rev{\textbf{48.3}} \\
\textit{n} & 100 & \textit{96} & \textit{76} & \textit{63} & \textit{41} \\
\bottomrule
\end{tabular}
\end{table}

\paragraph{Per-signature pass rates.} Table~\ref{tab:sig_pass} reports, for every signature, how often each model passed it (graded $1.0$) out of the sessions in which it was testable, with the pooled panel rate and its 95\% Wilson interval, and the random agent's rate for comparison. It makes the instability of Section~\ref{sec:study_gap} concrete, since many signatures pass on some repeats of a cell and not others. The same counts, with per-model intervals and effect sizes, are in \texttt{results/rebuttal\_v1/analysis/signature\_pass\_rates.csv} in the public repository.

\begin{table}[p]
\caption{Per-signature pass counts, $k/n$ = sessions graded $1.0$ out of sessions in which the signature was testable (at most ten per model). ``Panel'' pools the four models, with a 95\% Wilson interval.}
\label{tab:sig_pass}
\centering
\scriptsize
\resizebox{\textwidth}{!}{%
\begin{tabular}{@{}llccccll@{}}
\toprule
\textbf{Task} & \textbf{Signature} & \textbf{\texttt{gemini}} & \textbf{\texttt{kimi}} & \textbf{\texttt{nano}} & \textbf{\texttt{qwen}} & \textbf{Panel [95\% CI]} & \textbf{Floor} \\
\midrule
\texttt{random\_dot\_motion} & \texttt{coherence\_accuracy\_psychometric} & 4/10 & 1/9 & 0/3 & 0/2 & 0.21 [0.09, 0.40] & 0/10 \\
 & \texttt{coherence\_rt\_chronometric} & 0/10 & 0/7 & 0/3 & 0/1 & 0.00 [0.00, 0.15] & 1/10 \\
 & \texttt{above\_chance\_low\_coherence} & 5/10 & 0/8 & 0/3 & 0/1 & 0.23 [0.10, 0.43] & 1/10 \\
 & \texttt{above\_chance\_overall} & 9/10 & 0/10 & 0/7 & 0/7 & 0.26 [0.15, 0.43] & 2/10 \\
\midrule
\texttt{grid\_bandit} & \texttt{kraken\_avoidance\_in\_risky} & 9/10 & 4/5 & 4/10 & 2/8 & 0.58 [0.41, 0.73] & 0/10 \\
 & \texttt{value\_directed\_exploitation} & 7/10 & 1/4 & 4/10 & 7/10 & 0.56 [0.39, 0.71] & 1/10 \\
 & \texttt{above\_chance\_reward} & 10/10 & 5/5 & 2/10 & 3/9 & 0.59 [0.42, 0.74] & 0/10 \\
 & \texttt{within\_block\_learning} & 9/10 & 2/4 & 5/10 & 1/10 & 0.50 [0.34, 0.66] & 0/10 \\
 & \texttt{within\_block\_rt\_speedup} & 0/10 & 0/4 & 2/10 & 1/10 & 0.09 [0.03, 0.23] & 0/10 \\
\midrule
\texttt{marbles\_risk} & \texttt{ev\_sensitivity} & 10/10 & 5/8 & 1/5 & 0/0 & 0.70 [0.49, 0.84] & 1/10 \\
 & \texttt{above\_chance\_ev\_optimal} & 10/10 & 7/10 & 1/10 & 0/5 & 0.51 [0.36, 0.67] & 0/10 \\
 & \texttt{magnitude\_effect\_high\_payoff} & 9/10 & 4/8 & 1/5 & 0/0 & 0.61 [0.41, 0.78] & 0/10 \\
 & \texttt{ev\_difference\_rt\_effect} & 0/10 & 0/8 & 0/5 & 0/0 & 0.00 [0.00, 0.14] & 0/10 \\
\midrule
\texttt{repeated\_games} & \texttt{tit\_for\_tat\_reciprocity\_pd} & 0/0 & 0/0 & 0/1 & 0/0 & 0.00 [0.00, 0.79] & 0/10 \\
 & \texttt{above\_chance\_coordination\_bos} & 1/8 & 1/7 & 0/4 & 0/4 & 0.09 [0.02, 0.27] & 0/10 \\
 & \texttt{cooperation\_when\_opp\_cooperated} & 8/8 & 4/4 & 0/0 & 0/0 & 1.00 [0.76, 1.00] & 0/9 \\
 & \texttt{round\_rt\_speedup} & 5/10 & 1/10 & 2/4 & 1/5 & 0.31 [0.17, 0.49] & 1/10 \\
\midrule
\texttt{serial\_recall} & \texttt{above\_chance\_recall} & 10/10 & 2/7 & 3/10 & 4/4 & 0.61 [0.44, 0.76] & 0/10 \\
 & \texttt{similarity\_aids\_recall} & 1/10 & 0/6 & 0/10 & 0/4 & 0.03 [0.01, 0.17] & 0/10 \\
 & \texttt{high\_vs\_low\_sim\_advantage} & 1/10 & 0/6 & 2/10 & 0/4 & 0.10 [0.03, 0.26] & 0/10 \\
 & \texttt{similarity\_rt\_effect} & 1/10 & 0/6 & 0/10 & 1/4 & 0.07 [0.02, 0.21] & 2/10 \\
 & \texttt{primacy\_effect} & 0/10 & 0/6 & 2/10 & 0/4 & 0.07 [0.02, 0.21] & 0/10 \\
\midrule
\texttt{visual\_recognition} & \texttt{above\_chance\_discrimination} & 0/10 & 0/9 & 0/6 & 0/0 & 0.00 [0.00, 0.13] & 1/10 \\
 & \texttt{above\_chance\_accuracy} & 0/10 & 0/9 & 0/8 & 0/0 & 0.00 [0.00, 0.12] & 0/10 \\
 & \texttt{low\_false\_alarm\_rate} & 8/10 & 3/8 & 2/6 & 0/0 & 0.54 [0.35, 0.72] & 1/10 \\
\midrule
\texttt{effort\_foraging} & \texttt{travel\_cost\_residence\_effect} & 0/10 & 0/10 & 0/9 & 0/3 & 0.00 [0.00, 0.11] & 0/10 \\
 & \texttt{above\_chance\_mvt\_match} & 0/10 & 0/10 & 0/10 & 0/10 & 0.00 [0.00, 0.09] & 0/10 \\
 & \texttt{above\_chance\_total\_reward} & 0/10 & 0/9 & 1/9 & 0/3 & 0.03 [0.01, 0.16] & 0/10 \\
 & \texttt{harvest\_rt\_speedup} & 0/10 & 4/10 & 2/10 & 1/10 & 0.17 [0.09, 0.32] & 1/10 \\
\midrule
\texttt{tiny\_alchemy} & \texttt{above\_chance\_discovery} & 0/9 & 2/5 & 1/3 & 4/7 & 0.29 [0.15, 0.49] & 0/10 \\
 & \texttt{non\_redundant\_attempts} & 6/9 & 0/0 & 0/0 & 1/1 & 0.70 [0.40, 0.89] & 0/10 \\
\midrule
\texttt{moral\_machine} & \texttt{utilitarian\_preference} & 9/9 & 8/8 & 3/3 & 0/0 & 1.00 [0.84, 1.00] & 1/10 \\
 & \texttt{age\_preference\_save\_young} & 9/9 & 3/8 & 0/3 & 0/0 & 0.60 [0.39, 0.78] & 0/9 \\
 & \texttt{species\_preference\_save\_human} & 9/9 & 5/8 & 0/3 & 0/1 & 0.67 [0.45, 0.83] & 0/10 \\
 & \texttt{legality\_preference\_save\_lawful} & 3/8 & 0/6 & 0/3 & 0/0 & 0.18 [0.06, 0.41] & 1/6 \\
\midrule
\texttt{phishing\_detection} & \texttt{above\_chance\_discrimination} & 10/10 & 10/10 & 10/10 & 6/10 & 0.90 [0.77, 0.96] & 0/10 \\
 & \texttt{learning\_effect} & 0/10 & 0/10 & 0/2 & 0/0 & 0.00 [0.00, 0.15] & 1/10 \\
 & \texttt{above\_chance\_overall} & 10/10 & 10/10 & 10/10 & 4/10 & 0.85 [0.71, 0.93] & 0/10 \\
 & \texttt{practice\_rt\_speedup} & 5/10 & 1/10 & 0/10 & 3/10 & 0.23 [0.12, 0.38] & 0/10 \\
\bottomrule
\end{tabular}%
}
\end{table}

\rev{\paragraph{Models present in the pilot but not the panel.} \texttt{grok-4.1-fast} and \texttt{qwen3-vl-30b-instruct} each contributed ten single-session rows to the earlier pilot and no repeats here, for different reasons. The \texttt{grok-4.1-fast} model was delisted from OpenRouter after the pilot and cannot be re-run at any price, while \texttt{qwen3-vl-30b-instruct} remains available and was dropped for budget reasons. Their pilot rows are excluded from every panel aggregate in this paper, since one session per cell is the design this revision replaces. We record them here rather than dropping them silently. The delisting is also a reminder that a benchmark's reproducibility depends on model availability, which no benchmark controls. We plan to ask leaderboard submissions to use models that remain publicly callable, so that entries can be re-scored against later specifications.}

Two attempted models (\texttt{glm-4.6v}, \texttt{ui-tars-1.5-7b}) could not be scored at all, because Browser-Use rejected the format in which these models expressed their actions, at the first point in the experiment that required a response. These failures reflect incompatibility between model and scaffold rather than the models' abilities, and we excluded them.

\section{\rev{State-of-the-Art Model}}
\label{app:flagship}

\rev{The panel is economy-tier, so a low Level~3 could reflect the price of the models we chose rather than a limit of current multimodal models. To check, we ran \texttt{gpt-5.6-sol}, a state-of-the-art model at the time of writing, once on each of the ten tasks under the same scaffold and scoring. Table~\ref{tab:flagship} gives the per-task comparison against the panel's per-task means over measurable sessions.}

\rev{Three results are worth separating. The model beat the panel's per-task mean Level~3 on eight of ten tasks and gained most where the panel was weakest, reaching a perfect Level~3 on \texttt{visual\_recognition}, on which the panel averaged $0.235$. Against the panel's best member on each task it led on only four of ten, so it is not distinguishable from \texttt{gemini-3-flash-preview}, and the two comparisons must be read together. Its most consistent advantage was completion, with $L_1 = 1.000$ on nine of ten tasks against a panel mean of $0.750$ over all sessions. A large share of the composite gap therefore follows from finishing experiments that the economy panel abandons rather than from closer alignment. The \texttt{serial\_recall} task is the one on which it trailed the panel on completion ($0.600$ against $0.655$).}

\rev{These values come from a single session per task. The main study shows that Level~3 varies widely across repeated sessions for every panel model, and there is no reason to assume this model is exempt, so none of the per-task differences below should be read on its own. Repeats would be needed before this model could support a claim about any one task. We report it because it indicates how far the panel's results generalize, not because one session settles anything.}

\begin{table}[h]
\caption{\rev{State-of-the-art model against the panel, per task. Panel columns are means over the four models' measurable sessions on each task, and \texttt{gpt-5.6-sol} is a single session per task. Rows are ordered by Level~3 gain over the panel mean. The final row is the unweighted mean of the ten per-task values, and so differs slightly from the session-weighted panel means quoted in Section~\ref{sec:study_gap}.}}
\label{tab:flagship}
\centering
\small
\begin{tabular}{@{}lrrrr@{}}
\toprule
\textbf{Task} & \textbf{Panel L3} & \textbf{\texttt{sol} L3} & \textbf{Panel comp.} & \textbf{\texttt{sol} comp.} \\
\midrule
\texttt{visual\_recognition} & 0.235 & 1.000 & \rev{35.1} & \rev{99.1} \\
\texttt{effort\_foraging} & 0.166 & 0.636 & \rev{26.9} & \rev{77.7} \\
\texttt{moral\_machine} & 0.786 & 1.000 & \rev{83.3} & \rev{100.0} \\
\texttt{marbles\_risk} & 0.666 & 0.818 & \rev{71.3} & \rev{86.0} \\
\texttt{random\_dot\_motion} & 0.390 & 0.500 & \rev{34.1} & \rev{71.8} \\
\texttt{phishing\_detection} & \rev{0.672} & 0.773 & \rev{67.1} & \rev{77.5} \\
\texttt{repeated\_games} & 0.443 & 0.500 & \rev{53.5} & \rev{60.5} \\
\texttt{grid\_bandit} & 0.639 & 0.667 & \rev{52.9} & \rev{55.0} \\
\texttt{serial\_recall} & \rev{0.374} & 0.333 & \rev{38.0} & \rev{54.3} \\
\texttt{tiny\_alchemy} & 0.562 & 0.333 & \rev{67.3} & \rev{51.6} \\
\midrule
\textbf{Mean} & \rev{\textbf{0.493}} & \textbf{0.656} & \rev{\textbf{52.9}} & \rev{\textbf{73.3}} \\
\bottomrule
\end{tabular}
\end{table}

\section{Composite-Weight and Grading-Rule Sensitivity}
\label{app:weights}

To check that the ranking is not an artifact of the chosen composite weights $(0.15, 0.35, 0.50)$, Table~\ref{tab:weight_ablation} recomputes per-model composites under three alternatives. Per-cell L1, L2, and L3 values are unchanged. Only the weighting in Equation~\ref{eq:composite} differs across columns, \rev{with the Level~2 weight redistributed in proportion on the two tasks that have no Level~2,} and all columns average over all sessions (Section~\ref{sec:scoring}).

Averaged over all sessions, the ordering is identical under all four schemes, including the signature-only scheme $(0,0,1)$ that discards L1 and L2 entirely. Under the measurable-session convention, ranks two to four already depend on the convention (Section~\ref{sec:study}). There the three weighted schemes preserve the ordering, and the signature-only scheme moves \texttt{qwen3-vl-235b-instruct} from third to second. The \texttt{gemini-3-flash-preview} model leads under every scheme and convention. \rev{The largest shift arises when scores are averaged only over measurable sessions, on which \texttt{qwen3-vl-235b-instruct} rises from $26.4$ to $48.3$ over its 41 measurable sessions (Appendix~\ref{app:per_task}).} Coverage is reported alongside every composite so that the difference between the two conventions is visible rather than hidden in the aggregate.

\begin{table}[h]
\caption{\rev{Per-model mean composite under four weight schemes, averaged over all sessions. Headline weights are $(0.15, 0.35, 0.50)$. Coverage is the number of measurable sessions out of each model's one hundred.}}
\label{tab:weight_ablation}
\centering
\small
\begin{tabular}{@{}lccccc@{}}
\toprule
\textbf{Model} & \textbf{Headline} & \textbf{Signature-emph.} & \textbf{Balanced} & \textbf{Signature-only} & \textbf{Coverage} \\
& $(0.15, 0.35, 0.50)$ & $(0.10, 0.30, 0.60)$ & $(0.20, 0.40, 0.40)$ & $(0, 0, 1)$ & \\
\midrule
\texttt{gemini-3-flash-preview} & \rev{58.61} & \rev{56.80} & \rev{60.46} & \rev{54.37} & 96 \\
\texttt{kimi-k2.5} & \rev{44.23} & \rev{42.29} & \rev{46.21} & \rev{38.44} & 76 \\
\texttt{gpt-5.4-nano} & \rev{36.07} & \rev{33.87} & \rev{38.38} & \rev{30.36} & 63 \\
\texttt{qwen3-vl-235b-instruct} & \rev{26.41} & \rev{24.68} & \rev{28.24} & \rev{22.78} & 41 \\
\bottomrule
\end{tabular}
\end{table}

\paragraph{Grading rule.} The $\{0, 0.5, 1\}$ grade in Equation~\ref{eq:l3} discretizes a continuous test outcome, and a signature near the significance boundary moves a full step on a small change in $p$. We therefore regraded every testable signature observation under three alternatives, keeping a wrong-direction result at $0$ in each. The first uses six bins on the one-sided $p$, grading $1.0$ below $0.01$, then $0.8$, $0.6$, $0.4$, and $0.2$ below $0.05$, $0.10$, $0.20$, and $0.50$, and $0$ above. The second grades continuously in $p$ ($1-2p$, floored at $0$), and the third grades continuously in effect size ($2|e|$, capped at $1$). Session L3 and composite are recomputed from the stored L1 and L2, and the current rule reproduces every stored score exactly. The absolute Level~3 scale shifts, as it must, but the agent--floor gap moves by at most $0.021$ and the model ranking is identical under all four rules (Table~\ref{tab:grading_rule}), so no conclusion depends on the discretization. The regrade is \texttt{scripts/grading\_rule\_sensitivity.py} in the public repository.

\begin{table}[h]
\caption{Level~3 under four grading rules. Panel columns are means over the $276$ measurable sessions, and floor columns are means over all $100$ random-agent sessions. Ranking is by measurable-session composite, and is \texttt{gemini} $>$ \texttt{kimi} $>$ \texttt{qwen} $>$ \texttt{nano} under every rule.}
\label{tab:grading_rule}
\centering
\small
\begin{tabular}{@{}lccccl@{}}
\toprule
\textbf{Grading rule} & \textbf{Floor L3} & \textbf{Panel L3} & \textbf{L3 gap} & \textbf{Comp.\ gap} & \textbf{Ranking} \\
\midrule
Current $\{0, 0.5, 1\}$ & 0.184 & 0.490 & 0.306 & 20.71 & unchanged \\
Six bins on $p$ & 0.110 & 0.410 & 0.300 & 20.62 & unchanged \\
Continuous in $p$ & 0.157 & 0.458 & 0.302 & 20.72 & unchanged \\
Continuous in effect size & 0.072 & 0.357 & 0.285 & 19.99 & unchanged \\
\bottomrule
\end{tabular}
\end{table}

\clearpage


\end{document}